%% file: ani_paper.tex
\documentclass[]{fairmeta}
\usepackage{graphicx}
\usepackage{subcaption}
\usepackage{caption}
\usepackage{wrapfig}
\usepackage{multirow}
\usepackage{tcolorbox}   
\usepackage{subcaption}  
\usepackage{xcolor}
\tcbuselibrary{listings, breakable, skins}
\usepackage{amsmath}
\usepackage{xspace}
\usepackage{inconsolata}
\newcommand{\ROneLlamaSeventyB}{\texttt{R1{-}Llama{-}70B}\xspace}
\newcommand{\QwenThreeThirtyTwoB}{\texttt{Qwen3{-}32B}\xspace}
\newcommand{\behaviorlist}{behavior handbook\xspace}
\newcommand{\metacogexpert}{\texttt{Metacognitive Strategist}\xspace}
\newcommand{\teacher}{\texttt{Teacher}\xspace}
\newcommand{\student}{\texttt{Student}\xspace}

\title{Metacognitive Reuse: Turning Recurring LLM Reasoning Into Concise Behaviors}

\author[1,2]{Aniket Didolkar}
\author[1]{Nicolas Ballas}
\author[1, 3, \dagger]{Sanjeev Arora}
\author[1, \dagger]{Anirudh Goyal}

\affiliation[1]{Meta}
\affiliation[2]{Mila-Quebec AI Institute, University of Montreal}
\affiliation[3]{Princeton University}

\contribution[*]{Work done at Meta}
\contribution[\dagger]{Joint last author}

\abstract{Large language models (LLMs) now solve multi-step problems by emitting extended chains of thought. During the process, they often re-derive the same intermediate steps across problems, inflating token usage and latency. This saturation of the context window leaves less capacity for exploration. We study a simple mechanism that converts recurring reasoning fragments into concise, reusable “behaviors” (name + instruction) via the model’s own \emph{metacognitive analysis of prior traces}. These behaviors are stored in a “behavior handbook” which supplies them to the model in-context at inference or distills them into parameters via supervised fine-tuning. This approach achieves improved test-time reasoning across three different settings -   1) \textbf{Behavior-conditioned inference}: Providing the LLM relevant behaviors in-context during reasoning reduces number of reasoning tokens by up to 46\% while matching or improving baseline accuracy; 2) \textbf{Behavior-guided self-improvement}:  Without any parameter updates, the model improves its own future reasoning by leveraging behaviors from its own past problem solving attempts. This yields up to 10\% higher accuracy than a naive critique-and-revise baseline; and 3) \textbf{Behavior-conditioned SFT}: SFT on behavior-conditioned reasoning traces is more effective at converting non-reasoning models into reasoning models as compared to vanilla SFT. Together, these results indicate that turning slow derivations into fast procedural hints enables LLMs to remember how to reason, not just what to conclude. }

\date{\today}
\correspondence{\email{adidolkar123@gmail.com}, \email{agi@meta.com}}

\begin{document}

\maketitle

\section{Introduction}
\label{sec:intro}

LLMs have made rapid progress on mathematics, coding and other multi-step tasks by generating long, deliberative chains-of-thought \citep{wei2022chain, guo2025deepseek, shao2024deepseekmath, openai2024openaio1card, muennighoff2025s1, ye2025limo, gao2024designing,lambert2024tulu,team2025kimi}. Yet, this capability exposes a structural inefficiency: each new problem triggers reconstruction of ubiquitous sub-procedures (e.g., finite-series sums, case splits, unit conversions), inflating token usage and latency. For instance, suppose the LLM  derives the finite geometric series formula while solving one problem. Can it avoid re‑deriving from scratch when similar reasoning is needed for another problem? Current inference loops lack a mechanism to promote frequently rediscovered patterns into a compact, retrievable form. 

We introduce a \emph{metacognitive pathway} that extracts and reuses such patterns. Given a problem, the model first solves it, then reflects on its trace to identify generalizable steps, and finally emits a set of behaviors—short, actionable instructions with canonical names. These behaviors populate a searchable handbook (a procedural memory) that can be provided in-context at test time or internalized through supervised fine-tuning. This provides a framework for turning verbose derivations into \textit{quick reflexes}.

Unlike typical memory/Retrieval-Augmented Generation (RAG) systems that store declarative facts, the handbook targets procedural knowledge \citep{WillinghamNissenBullemer1989ProceduralKnowledge} about how to think. This procedural memory contrasts sharply with most existing “memory” add‑ons for LLMs, including RAG, which target declarative knowledge for tasks such as factual question-answering \citep{borgeaud2022improving, lewis2020retrieval}. Instead of being assembled from curated documents or knowledge graphs—rather, it is generated by the model itself. It emerges from the model’s own metacognitive cycle: critiquing its own chain‑of‑thought and abstracting repeated reasoning patterns into behaviors. 

We evaluate three instantiations of the proposed framework. (i) Behavior-conditioned inference: Providing behaviors obtained by solving  questions in-context results in reasoning chains that utilize up to 46\% fewer tokens while improving or maintaining strong performance across MATH and AIME benchmarks (ii) Behavior-guided self-improvement: While solving a problem, providing the model access to behaviors extracted by itself from its own past attempts for that question improves accuracy by up to 10\% compared to naive self-improvement baseline. (iii) Behavior-conditioned SFT: training on reasoning traces generated via behavior-conditioned inference yields models that are both more accurate and more concise than models trained on ordinary traces, especially when turning non-reasoning models into reasoning models.

\begin{figure}
    \centering
    \includegraphics[width=\linewidth]{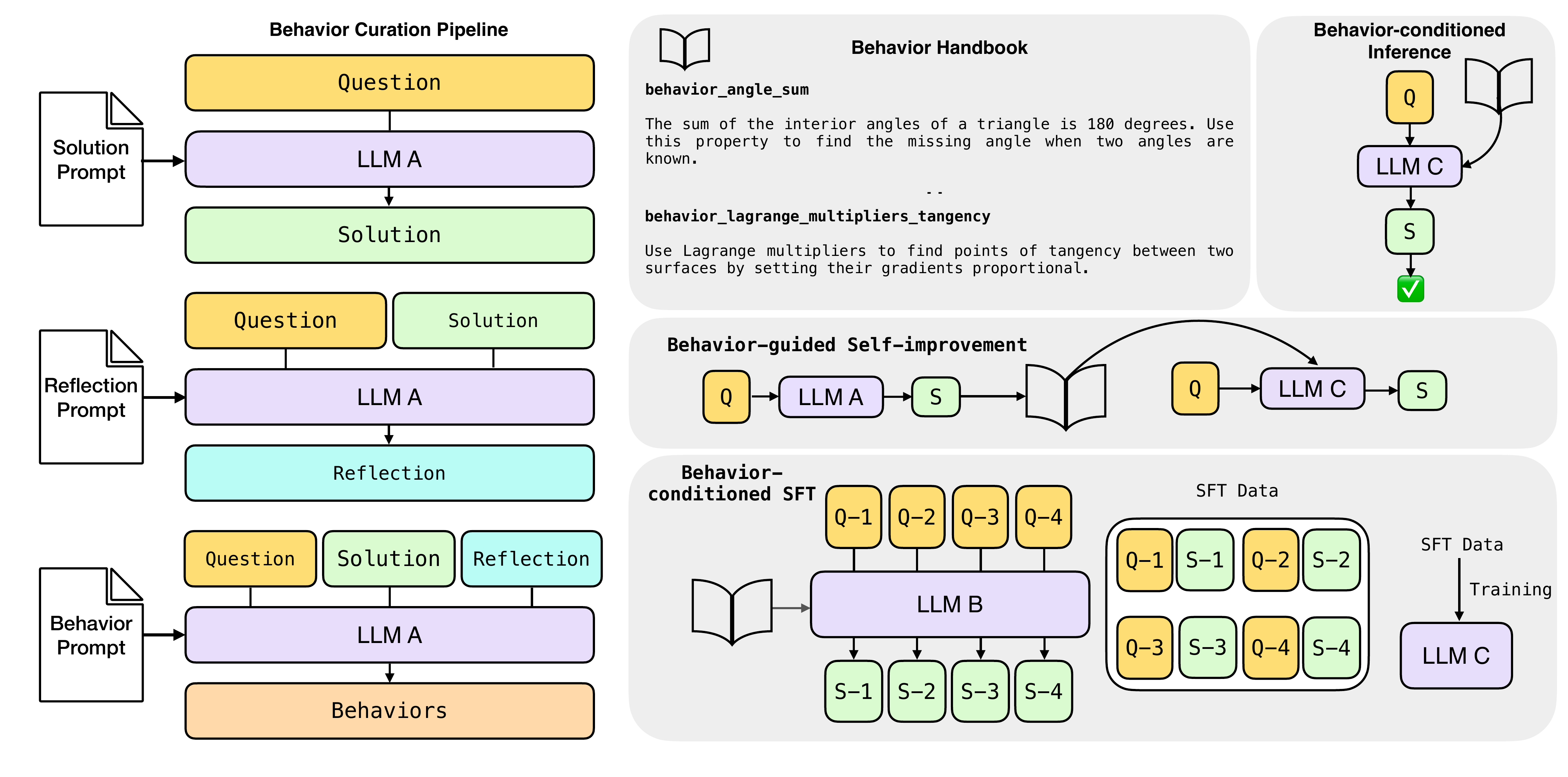}
    \caption{\textbf{Behavior Curation Pipeline (left)}: All the 3 stages of behavior curation pipeline are shown. The stages are described in detail in Section \ref{sec:thought_pipeline} with detailed prompts for each stage shown in Figure \ref{fig:behavior_prompts}. LLM A refers to the \metacogexpert. \textbf{Reasoning with behaviors (right)}: This part showcases various ways in which behaviors are utilized for LLM reasoning. For behavior-conditioned inference and behavior-guided self-improvement, behaviors retrieved from the \behaviorlist, are provided in-context to the \student model (LLM C) during inference. For behavior-conditioned SFT, a training dataset is created using a \teacher LLM (LLM B) via behavior-conditioned inference and then a \student LLM is trained on the resulting (question, response) pairs. After training, during inference, the \student LLM is queried with the given question without any behaviors in-context.}
    \label{fig:behavior_demo}
    \vspace{-3mm}
\end{figure}

Contributions.
\begin{enumerate}
    \item We formalize behaviors as named, reusable reasoning instructions discovered by metacognitive reflection over solution traces.
    \item We introduce a three-step approach to employ an LLM to extract behaviors from its own reasoning traces  (Section \ref{sec:thought_pipeline}).
    \item We develop three settings for utilizing these behaviors: behavior-conditioned inference, behavior-guided self-improvement, and behavior-conditioned SFT. (Section \ref{sec:behaviors_for_reasoning})
    \item We provide empirical evidence of the effectiveness of our behavior-based approach across each of the three settings, evaluated on challenging mathematical benchmarks such as MATH \citep{hendrycks2021measuring} and AIME\textendash24/25 \citep{MAA_AIME_2024, MAA_AIME_2025} (Section~\ref{sec:experiments}).
    \item We discuss some limitations and challenges of the proposed framework—e.g., lack of dynamic behavior retrieval during long solutions, usage of behaviors beyond math, constructing large scale behavior handbooks across various domains, etc. 
\end{enumerate}

By converting frequently rediscovered steps into compact procedures, behavior handbooks encourage LLMs to remember how to think. This simple addition to the reasoning stack improves token efficiency and suggests a path toward models that accumulate procedural knowledge over time. 

\subsection{Paper Outline}
The paper starts with describing the pipeline for curating the behaviors (Section \ref{sec:thought_pipeline}) followed by various ways in which behaviors are utilized for improved reasoning (Section \ref{sec:behaviors_for_reasoning}). The experiment section describes the corresponding experiment results. Section \ref{sec:bci} presents results for behavior-conditioned inference on the MATH \citep{hendrycks2021measuring} and the AIME datasets \citep{MAA_AIME_2024, MAA_AIME_2025} showing that the proposed approach exhibits similar or improved performance compared to normal inference while reducing token usage by up to 46\%. Section \ref{sec:self-improvement} presents the self-improvement experiment where the behavior-guided approach uses behaviors as \textit{lessons} for scalable self-improvement achieving up to 10\% higher accuracy as compared to the baseline self-improvement approach at the highest considered token budget of 16{,}384. Finally, the SFT experiments (Section \ref{sec:sft}) show that using behavior-conditioned inference to generate reasoning traces for SFT results in stronger reasoning models as compared to performing SFT with vanilla reasoning traces. 


\section{Related Work}

\paragraph{Efficient Reasoning with LLMs} Reinforcement‑tuned, long‑form chain‑of‑thought (CoT) prompting has enabled recent LLMs to tackle highly complex problems in mathematics, logic, and code \citep{openai2024openaio1card,guo2025deepseek,shao2024deepseekmath,gao2024designing,lambert2024tulu,team2025kimi,zeng2025simplerl,song2025fastcurl,muennighoff2025s1,ye2025limo}. Although CoT lets a model “think out loud” for seconds or minutes, a growing literature seeks to shorten those traces while preserving accuracy. Skeleton‑of‑Thought first drafts an outline and then expands each bullet in parallel \citep{ning2023skeleton}; TokenSkip trains models to omit redundant tokens altogether \citep{xia2025tokenskip}; Dynasor inserts early‑exit probes that halt generation once successive probes agree on the answer \citep{fu2024efficiently}; and MinD constrains the model to concise, single‑trajectory blocks across multiple turns \citep{zeng2025done}. The proposed approach shares the efficiency goal but diverges in two ways: (i) we do not explicitly train the model to be terse—efficiency emerges after the model abstracts recurring reasoning fragments into reusable behaviors; and (ii) these behaviors also improve solution quality, as shown in the SFT experiment (Section~\ref{sec:sft}), where training on behavior‑conditioned traces outperforms training on long-form extended chain-of-thought traces.

\paragraph{Metacognitive abilities of LLMs} 
Metacognition  refers to humans' ``thinking about thinking'' \citep{flavell1979metacognition}. \citet{didolkar2024metacognition} suggested that one LLM analog of metacognition is the ability to  extract reusable ``skills'' from the CoT of LLMs, and showed that frontier LLMs can extract meaningful skill catalogs from task datasets. 
Such LLM-extracted skill catalogs were
used to create more difficult math questions in \citet{shah2025aiassistedgenerationdifficultmath}, by requiring questions to involve  skill compositions. 
\citet{kaur2025instructskillmix} follow a similar approach for instruction-following skills. \citet{he2025adaptmiadaptiveskillbasedincontext} use skill categories to study in-context learning in smaller language models. 
The novelty of our work is to apply metacognitive thinking to help reasoning models with their longer and complicated reasoning traces.

\paragraph{Memory in LLMs} 
Current memory implementations for LLM mainly rely on an external store of factual knowledge (such as Wikipedia) that the model can search \citep{borgeaud2022improving, lewis2020retrieval, guu2020retrieval, shi2023replug,he2021efficient}. Retrieval‑augmented generation pulls passages from this factual memory at inference time and conditions the decoder on the evidence to answer knowledge‑intensive queries. More recently, retrieval has been woven directly into multi‑step reasoning, with methods like ReAct and IR‑CoT interleaving “think$\rightarrow$retrieve$\rightarrow$think” loops to reduce hallucinations \citep{yao2023react,trivedi2022interleaving}. These implementations of memory mainly store declarative knowledge which corresponds to \textit{what} is true, not \textit{how to} think. Procedural knowledge—skills and routines acquired through repetition—remains largely unexplored. Our proposed \behaviorlist is one instantiation of such procedural memory for LLMs: it captures how‑to strategies distilled from repeated reasoning patterns and stores them for future reuse.

\section{Behaviors from Reasoning traces} \label{sec:thought_pipeline}

Reasoning LLMs emit a long chain-of-thought (CoT) which we will also refer to as a reasoning trace. We define a \emph{behavior} as a reusable skill---a concise piece of knowledge---distilled from an LLM's chain of thought. 
Such behaviors can be invoked across tasks to make inference-time reasoning both faster and more accurate. 
Each behavior is represented as a \texttt{(name, instruction)} pair. 
For example:
\begin{center}
\[
\texttt{systematic\_counting}\;\rightarrow\;
\begin{aligned}[t]
  &\text{Systematically count possibilities by examining each digit’s contribution}\\
  &\text{without overlap; this prevents missed cases and double-counts.}
\end{aligned}
\]
\end{center}
The remainder of this section describes the process of deriving behaviors from LLM-generated reasoning traces.

Figure \ref{fig:behavior_demo} depicts the entire pipeline. The framework employs LLMs in 3 different roles - 1) A \metacogexpert (LLM A) which extracts behaviors from its own reasoning traces; 2) A \teacher (LLM B) which generates data for SFT training; and 3) A \student (LLM C) whose reasoning is aided by behaviors either via behavior-conditioned inference or behavior-conditioned SFT. We dive deeper into each of these roles in the following sections. First, we describe the working of the \metacogexpert.

\begin{figure}
    \centering
    \includegraphics[width=\linewidth, trim={0 10cm 0 0},clip]{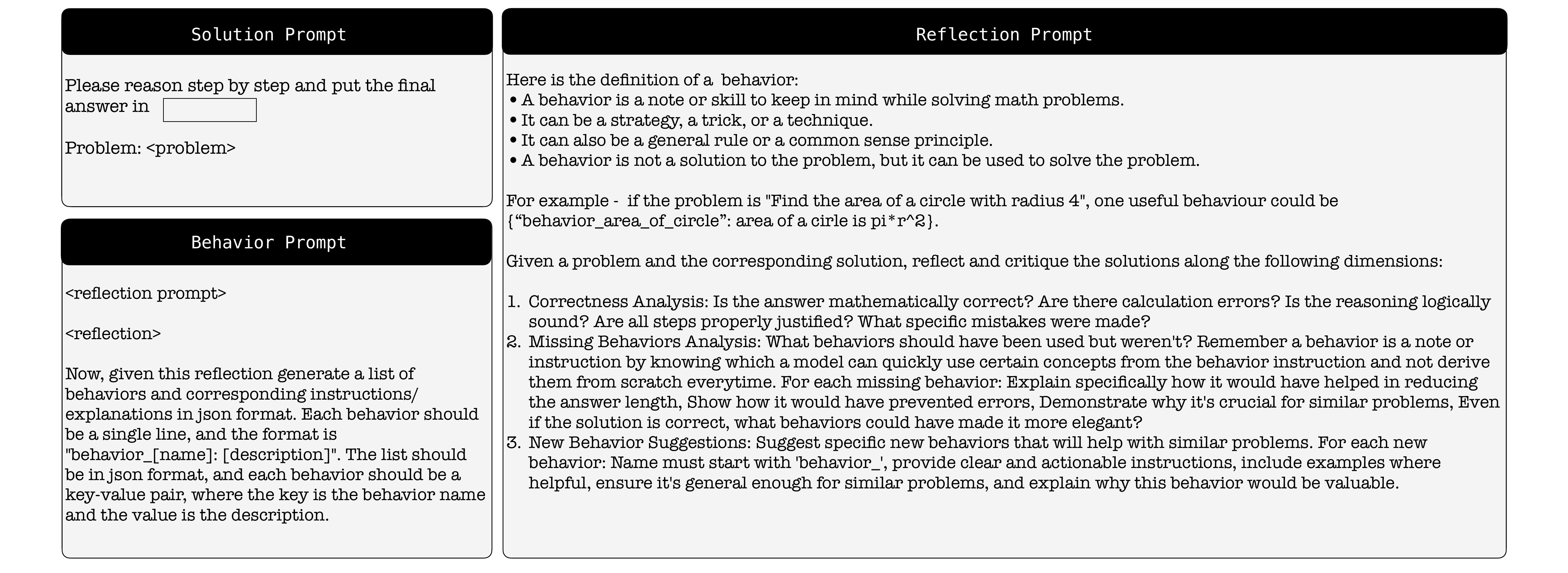}
    \caption{Prompts used for extracting behaviors from solutions are listed in this Figure. \textbf{Solution Prompt} is used to map the questions to solutions containing reasoning traces. Next, the \textbf{Reflection Prompt} is employed to generate a reflection for the solution followed by using the \textbf{Behavior Prompt} to generate the behaviors.}
    \label{fig:behavior_prompts}
\end{figure}

\paragraph{Extracting Behaviors} To extract behaviors, the \metacogexpert produces a solution for a given question which consists of the reasoning trace + the final answer. The prompt for this interaction is shown in Figure \ref{fig:behavior_prompts} (\texttt{Solution Prompt}). The question–solution pair is then fed to the \metacogexpert again to generate a \textit{reflection} which evaluates whether the reasoning is logically sound, the answer correct, and whether any new, reusable behaviors can be distilled to streamline future problem solving (See \texttt{Reflection Prompt} in Figure \ref{fig:behavior_prompts}). Finally, via another query the \metacogexpert converts the question, solution, and reflection into a set of \texttt{(name, instruction)} behavior entries, which are appended to an ever-growing \behaviorlist (See \texttt{Behavior Prompt} in Figure \ref{fig:behavior_prompts}). The \behaviorlist panel in Figure \ref{fig:behavior_demo} shows two example behaviors derived from the MATH \citep{hendrycks2021measuring} and AIME\textendash24 datasets \citep{MAA_AIME_2024} datasets. 

The \texttt{DeepSeek{-}R1{-}Distill{-}Llama{-}70B} (\ROneLlamaSeventyB) \citep{guo2025deepseek} is used as the \metacogexpert. 

\section{Behavior-Guided Reasoning} \label{sec:behaviors_for_reasoning}

This section discusses various ways in which the \behaviorlist is utilized for scalable and efficient reasoning.

\subsection{Behavior-conditioned inference}

One straightforward way to utilize the behaviors is providing a \student LLM access to those behaviors in-context during reasoning as shown in Figure \ref{fig:behavior_demo}. We term this as behavior-conditioned inference (BCI). Given a question $Q_i$, the proposed approach first retrieves relevant behaviors $B_i$ from the \behaviorlist. The behaviors, their corresponding instruction, and the question are then fed into the LLM to produce a solution-

\begin{align}
    (B_i, Q_i) \rightarrow S_i
\end{align}

\begin{wrapfigure}{r}{0.5\linewidth}   
  \vspace{-1.0\baselineskip}            
  \centering
  \includegraphics[width=\linewidth,trim={12cm 15cm 12cm 0},clip]%
                  {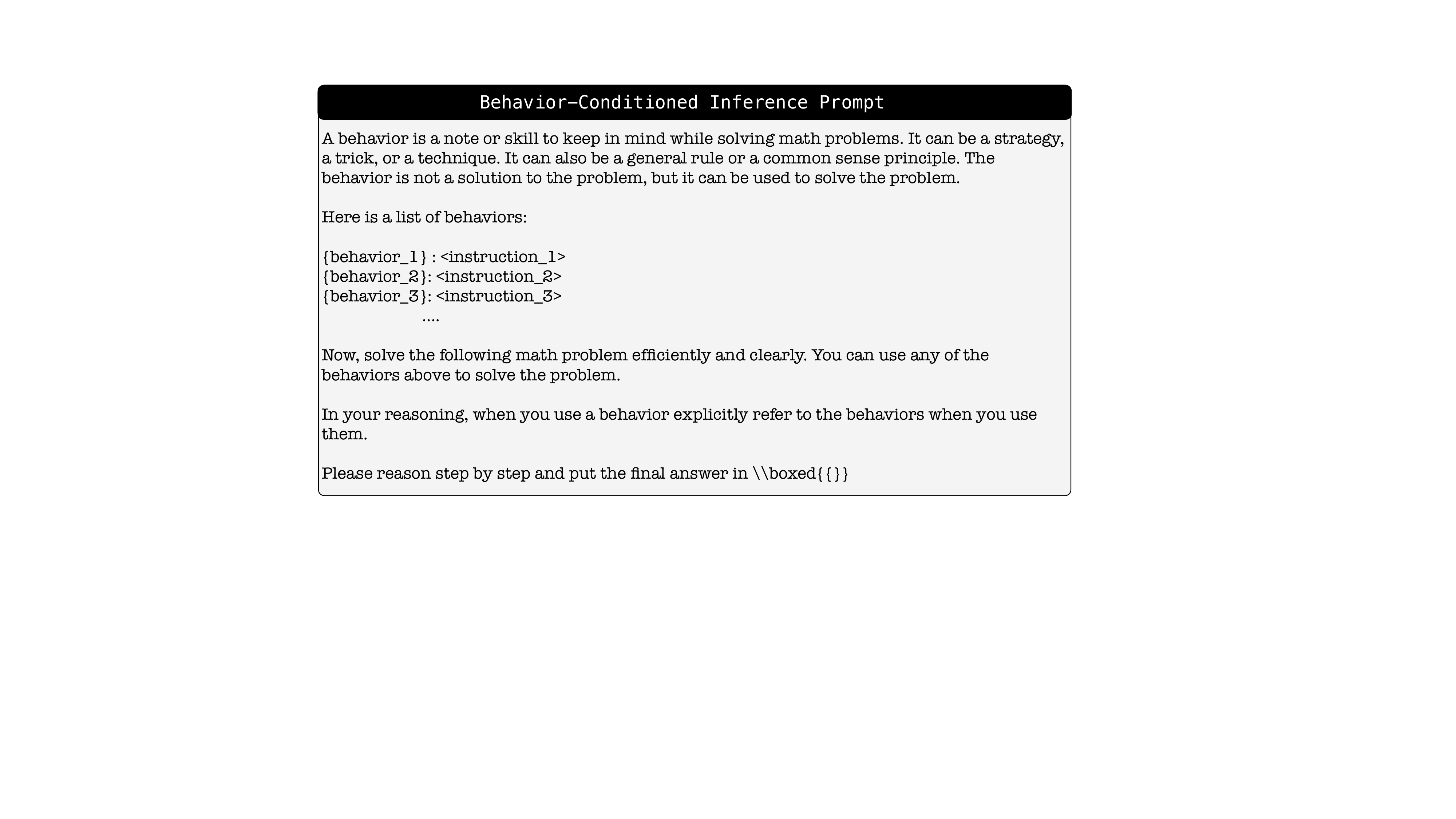}
  \caption{Prompt used for behavior‑conditioned inference (BCI).}
  \label{fig:behavior_conditioned_inference_prompt}
  \vspace{-0.5\baselineskip}            
\end{wrapfigure}

The exact prompt used for BCI is mentioned in Figure \ref{fig:behavior_conditioned_inference_prompt}. The form of the retrieval function, which retrieves relevant behaviors from the \behaviorlist for a given question, depends on the exact use-case. For instance, in the MATH dataset \citep{hendrycks2021measuring}, the retrieval function is based on topic-matching - for a question from a given topic, behaviors from that topic are retrieved. This is possible for the MATH dataset since all training and test set questions are annotated with one of 7 topics. Therefore, behaviors in the \behaviorlist  can be categorized using the topics of the questions that they were obtained from. Such retrieval is not possible for other datasets like  . AIME\textendash24, 25 \citep{MAA_AIME_2024, MAA_AIME_2025}. In that case, embedding-based retrieval is used for retrieving relevant behaviors. For a given question, the top-K behaviors ranked by cosine similarity in embedding space are selected. More details of this retrieval function are provided in Section \ref{sec:bci}. More information regarding the form of the retrieval functions used for each experiment is deferred to the Section \ref{sec:experiments} and described along with the corresponding experiment.

\subsection{Behavior-guided self-improvement}

Self-improvement is a given model's ability to improve its own reasoning. 
To achieve this, behaviors curated by a model from the reasoning traces of a particular question are then fed back into the model in-context to serve as lessons or hints to solve the same question or new questions.  The implementation closely follows  that of BCI and uses the same prompt. This process is depicted in Figure \ref{fig:behavior_conditioned_inference_prompt}. The \student LLM is the same as the \metacogexpert.

\subsection{Behavior-conditioned supervised fine-tuning}
Behavior‐conditioned inference still incurs a retrieval step and extra prompt tokens at test time to remind the model which behaviors to use. We eliminate this overhead by \emph{internalising} the behaviors through fine-tuning the given model on data generated via BCI. We term this as Behavior-conditioned supervised fine-tuning (BC\textnormal{-}SFT). The \metacogexpert generates the behaviors, the \teacher generates data using BCI, and the \student is fine-tuned on that data. The procedure is as follows:

\begin{enumerate}
    \item The \metacogexpert extracts behaviors for each question using the pipeline in Section \ref{sec:thought_pipeline} followed by the \teacher which generates a behavior‑conditioned response for each question using BCI.
    \item The \student model is fine-tuned on the resulting \emph{(question, behavior‑conditioned response)} pairs.
\end{enumerate}

This pipeline is depicted in Figure \ref{fig:behavior_demo} and more details are elaborated in Section \ref{sec:sft}. The \student model no longer needs behavior prompts; it spontaneously invokes the learned behaviors. This distillation setup converts the teacher’s deliberate, behavior‑annotated reasoning into a student's fast, intuitive, low‑token responses. Such a setup also allows us to evaluate the effectiveness of behavior-conditioned reasoning to equip a non-reasoning model with reasoning capabilities.

\begin{figure}[htbp]
  \centering
  
  \begin{subfigure}[b]{0.45\textwidth}
    \centering
    \includegraphics[width=\textwidth]{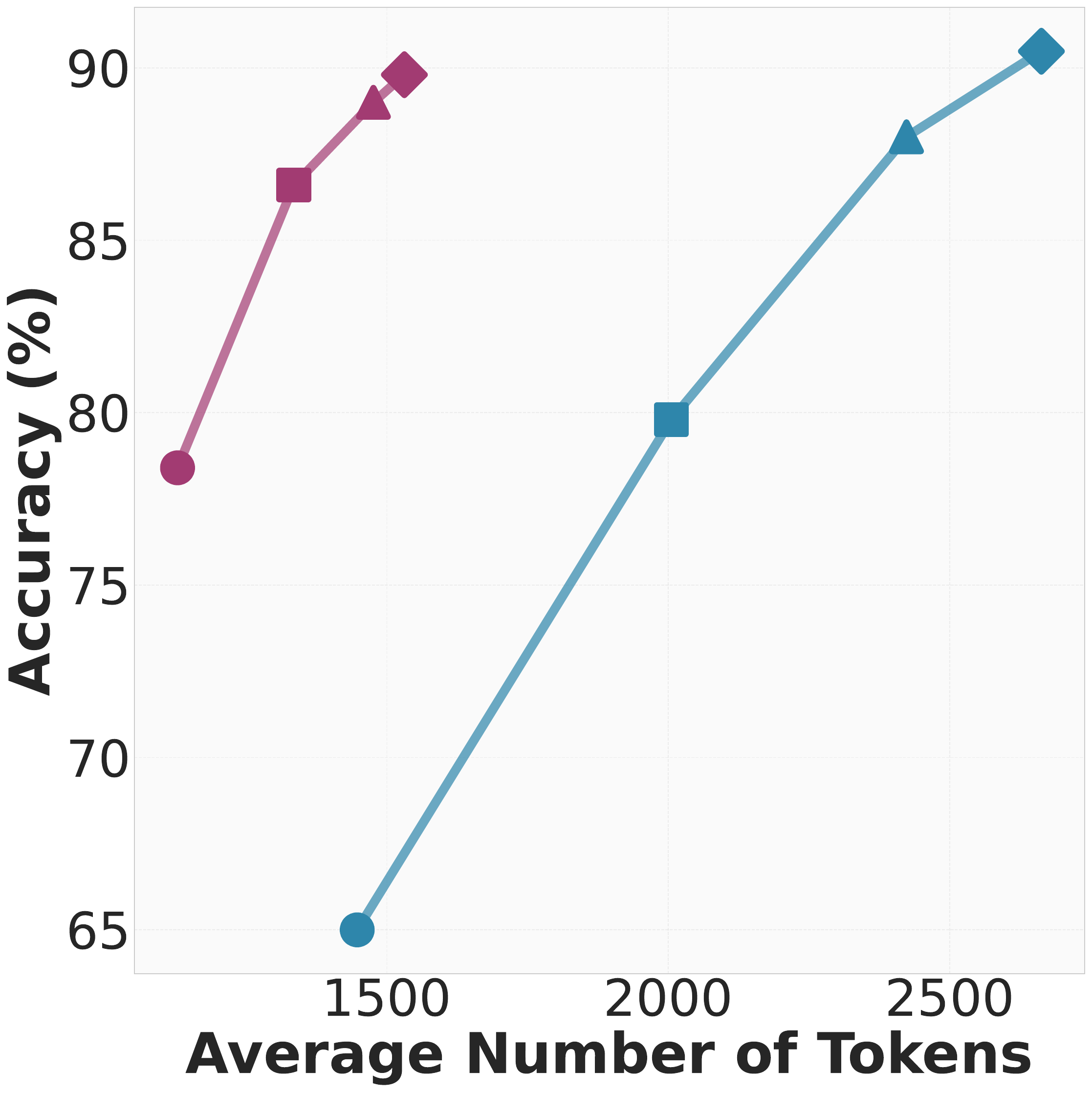}
    \caption{Behavior-conditioned Inference (BCI) on \ROneLlamaSeventyB}
    \label{fig:subfig1}
  \end{subfigure}
  \hfill
  \begin{subfigure}[b]{0.45\textwidth}
    \centering
    \includegraphics[width=\textwidth]{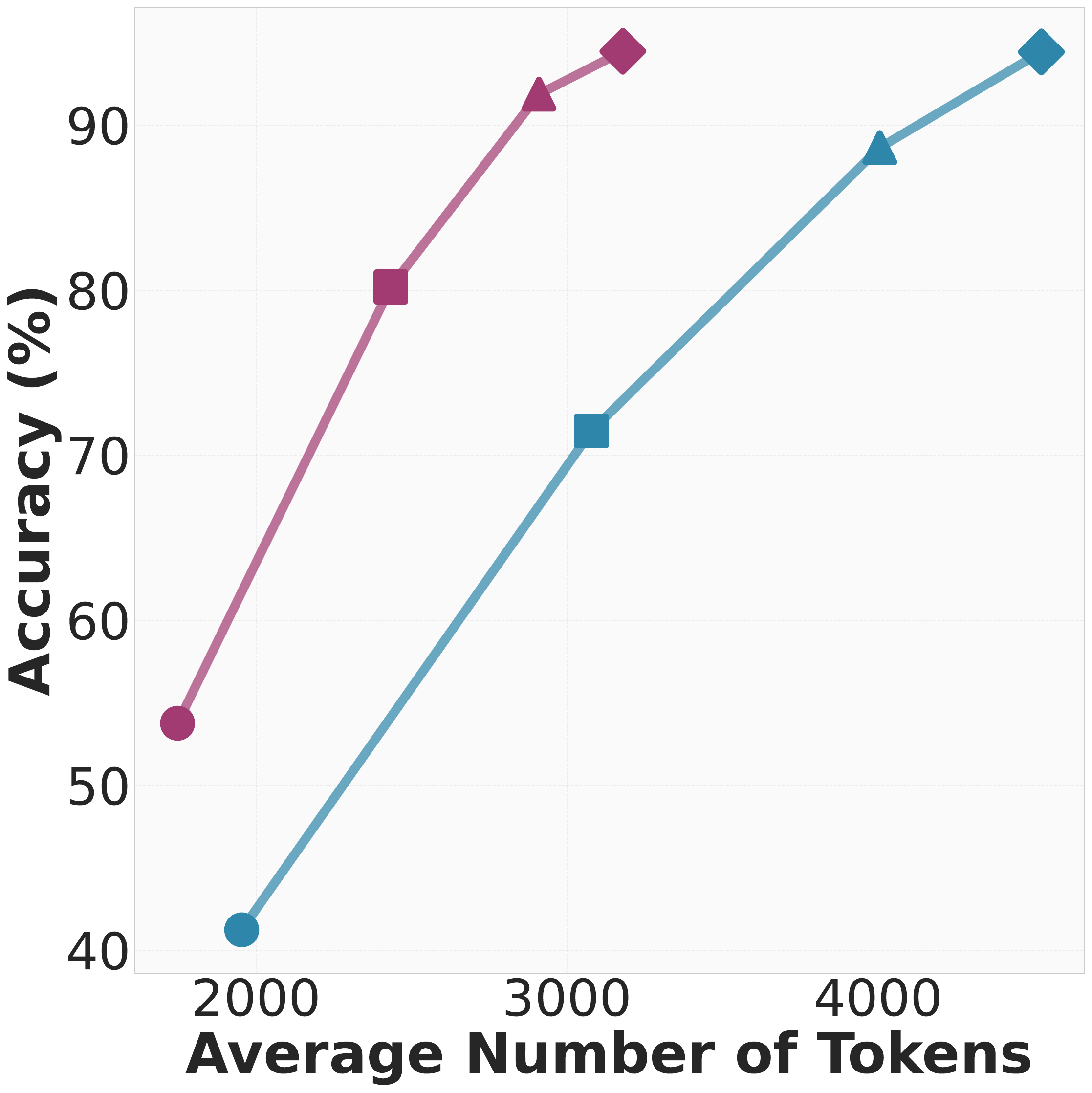}
    \caption{Behavior-conditioned Inference (BCI) on \QwenThreeThirtyTwoB}
    \label{fig:subfig2}
  \end{subfigure}
  \vspace{2mm}
  \includegraphics[width=\textwidth]{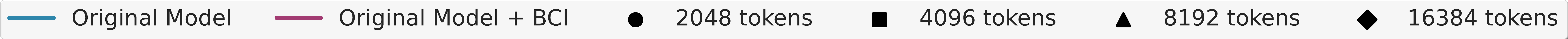}
  
  \caption{\textbf{Behavior-conditioned Inference (BCI) for MATH}. Using behaviors from \ROneLlamaSeventyB, BCI is applied on two models - \ROneLlamaSeventyB and \QwenThreeThirtyTwoB - while evaluating on the MATH-500 dataset \citep{hendrycks2021measuring}. The x-axis shows the average number of tokens produced per solution, the y-axis depicts the accuracy, and each point on the line corresponds to a particular hard token budget which is enforced during inference as shown in the legend. BCI achieves superior token efficiency producing answers with similar or improved accuracy while utilizing far fewer tokens than the base models.}
  \label{fig:math_comparison}
\end{figure}

\begin{table}[htbp]
\centering
\renewcommand{\arraystretch}{1.2}
\begin{tabular}{| >{\scriptsize}m{0.3\textwidth} 
                | >{\scriptsize}m{0.3\textwidth} 
                | >{\scriptsize}m{0.3\textwidth} |}
\hline
Two fair, 6-sided dice are thrown. What is the probability that the product of the two numbers is a multiple of 5?  Express your answer as a common fraction. &
Remmy wants to divide $10$ by $\frac{2}{3}$, but he cannot remember how to do that. By what number should he multiply $10$ to get the answer?&
Hillary has eleven coins, all dimes and nickels. In total, the coins are worth 75 cents. How many nickels does she have? \\
\hline
Alright, so I need to find the probability that the product of two fair 6-sided dice is a multiple of 5. Hmm, okay.

...

...

That would be $6$ (for the first die) multiplied by $6$ (for the second die), which gives us $36$ total possible outcomes. I think this is an application of the \textcolor{blue}{\texttt{behavior\_total\_outcomes}} behavior.

...

...

There are two scenarios where the product is a multiple of 5:
\begin{enumerate}\setlength\itemsep{1pt}
    \item The first die is a 5, regardless of what the second die shows.
    \item The second die is a 5, regardless of what the first die shows.
\end{enumerate}

Wait a second — if I just add these two scenarios together, I might be double-counting the case where both dice are 5. So, I need to adjust for that overlap to avoid counting it twice.

This sounds like the \textcolor{blue}{\texttt{behavior\_inclusion\_exclusion}} principle, where I add the probabilities of each event and then subtract the probability of their intersection....
 &
Alright, let me try to figure out how to solve this problem. So, Remmy wants to divide 10 by 2/3, but he's forgotten how to do that

...

...

As per the behavior \textcolor{blue}{\texttt{behavior\_reciprocal}}, I know that dividing by a fraction is the same as multiplying by its reciprocal. 

...

...

So, my initial answer of 15 seems correct.

Another way to think about it is using the behavior \textcolor{blue}{\texttt{behavior\_estimate\_reasonableness}}. If I estimate, dividing 10 by a number less than 1 should give me a number larger than 10. Since 2/3 is about 0.666, dividing 10 by 0.666 should be around 15, which matches my previous result....
&
Alright, so Hillary has eleven coins, and they're all dimes and nickels. The total value is 75 cents. I need to figure out how many nickels she has. Let me break this down step by step.

First, I'll use the behavior \textcolor{blue}{\texttt{behavior\_translate\_verbal\_to\_equation}} to convert the problem into mathematical equations. Let's let \( d \) represent the number of dimes and \( n \) represent the number of nickels.

I know two things: the total number of coins and the total value. So, I can write two equations. The first equation is about the number of coins:
\[
d + n = 11
\]
This is because she has eleven coins in total, and they're all dimes and nickels.

The second equation is about the total value. Since each dime is worth 10 cents and each nickel is worth 5 cents, the total value equation is:
\[
10d + 5n = 75
\].... \\
\hline
\end{tabular}
\caption{\textbf{Behavior-conditioned Inference examples} This table illustrates how \ROneLlamaSeventyB uses behaviors to ease problem solving. The behaviors used are highlighted in \textcolor{blue}{blue}.}
\label{tab:bci_examples}
\end{table}

\section{Experimental Results} \label{sec:experiments}

This section presents experimental results for each use-case described in section \ref{sec:behaviors_for_reasoning}. Unless otherwise specified, the decoding temperature is set to $0.6$ and top{\textendash}p is set to 0.95.

\subsection{Behavior-conditioned Inference for efficient Mathematical Reasoning} \label{sec:bci}

For the first use-case, BCI is applied to the MATH \citep{hendrycks2021measuring} and AIME\textendash24/25 \citep{MAA_AIME_2024, MAA_AIME_2025} datasets. Two open-source reasoning LLMs - DeepSeek-R1-Distill-Llama-70B (\ROneLlamaSeventyB) \citep{guo2025deepseek} and Qwen3-32B (\QwenThreeThirtyTwoB) \citep{bai2023qwen} - are used as \student candidates. 

\ROneLlamaSeventyB is employed as the \metacogexpert. More dataset-specific details are provided in the following paragraphs.

\paragraph{MATH Dataset} The \behaviorlist is curated using a random sample of 1k questions from the MATH training set. The reasoning traces, reflections, and the behaviors are generated with a token budget of 8{,}192. All reasoning traces are used for curating the \behaviorlist regardless of correctness. Here is the obtained distribution of behaviors per MATH subject - Algebra: 113, Prealgebra: 144, Counting: 128, Intermediate Algebra: 107, Geometry: 95, Number Theory: 107, and Precalculus: 91. Appendix Table \ref{tab:math_behavior}  presents a few example behaviors for each subject. 

\paragraph{Baseline} The baseline in this case runs normal inference on the same LLM by simply prompting it with the given problem and asking it to reason and produce the final answer in boxed format.

During inference, relevant behaviors are retrieved from the \behaviorlist for a given question based on topic-matching. The evaluation results on the MATH-500 set are presented in Figure \ref{fig:math_comparison}. The mean accuracy across 5 seeds is reported in the plot. The proposed BCI approach, achieves similar or improved accuracy while utilizing fewer tokens than the original model. Secondly, the performance still scales with increasing token-budget thus the proposed approach does not affect the model's existing capabilities in unwanted ways. 

A few example reasoning traces output by the \ROneLlamaSeventyB during BCI are presented in Table \ref{tab:bci_examples}. Only those parts of the trace which utilize behaviors are shown. Some behaviors encode core mathematical concepts such as \textcolor{blue}{\texttt{behavior\_total\_outcomes}} and \textcolor{blue}{\texttt{behavior\_inclusion\_exclusion}} while others encode more general problem solving strategies such as \textcolor{blue}{\texttt{behavior\_estimate\_reasonableness}} and \textcolor{blue}{\texttt{behavior\_translate\_verbal\_to\_equation}}.

\paragraph{AIME Datasets} Next, BCI is evaluated on the AIME\textendash24 \citep{MAA_AIME_2024} and AIME\textendash25 \citep{MAA_AIME_2025} datasets. The \behaviorlist for these datasets is created using the past versions of the AIME exams - specifically the AIME\textendash22 and AIME\textendash23 sets which have 30 questions each. \ROneLlamaSeventyB is again used as the \metacogexpert. Behaviors are generated for each of the 60 questions while generating 16 reasoning traces per question at a token budget of 16{,}384. The reflections and the behaviors are also generated at a token budget of 16{,}384 tokens. The final \behaviorlist consists of 1457 behaviors from 60 questions. Some examples of these behaviors are presented in the Appendix table \ref{tab:aime2223_behaviors}.

During inference, embedding based retrieval is used to fetch relevant behaviors per question. The BGE-M3 sentence transformer model \citep{chen2024bge} is used to encode both the query question text and all 1457 behaviors (and their corresponding instructions) from the AIME\textendash22/23 corpus into dense vector representations. A FAISS index \citep{douze2024faiss} is constructed over the behavior embeddings from the \behaviorlist from which the top-$k$ behaviors are retrieved for each question. In this case, $k$ is set to 40. Critically, the FAISS-based retrieval layer is scalable: in principle a very large, continually expanding, cross-domain library of behaviors can be maintained and only the most relevant behaviors to a given query can be retrieved from this library with manageable latency and memory cost. 

\begin{figure}[htbp]
  \centering
  
  \begin{subfigure}[b]{0.2\textwidth}
    \centering
    \includegraphics[width=\textwidth]{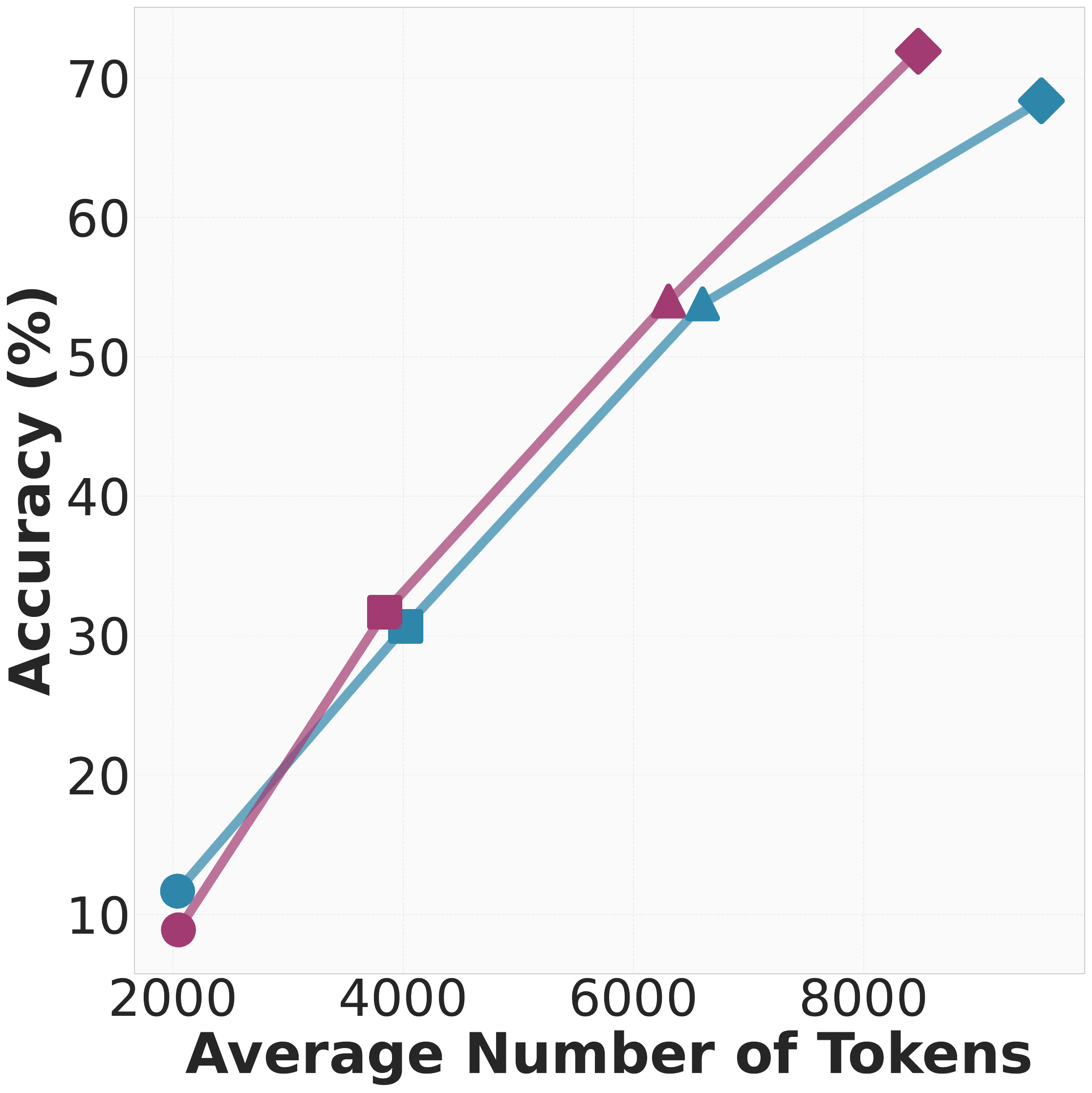}
    \caption{\ROneLlamaSeventyB}
    \label{fig:subfig2}
  \end{subfigure}
   \hfill
  \begin{subfigure}[b]{0.2\textwidth}
    \centering
    \includegraphics[width=\textwidth]{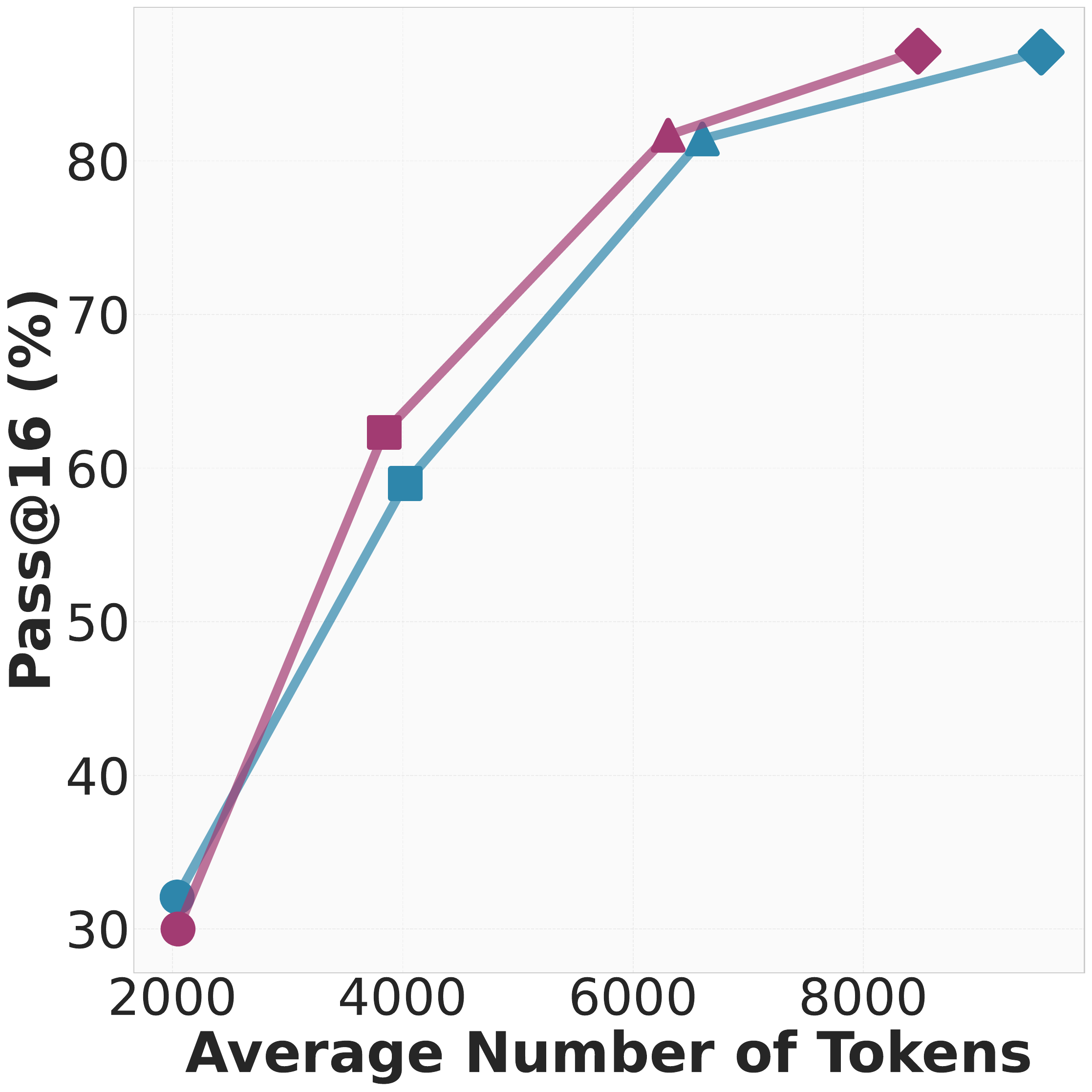}
    \caption{\ROneLlamaSeventyB}
    \label{fig:subfig2}
  \end{subfigure}
    \hfill
  \begin{subfigure}[b]{0.2\textwidth}
    \centering
    \includegraphics[width=\textwidth]{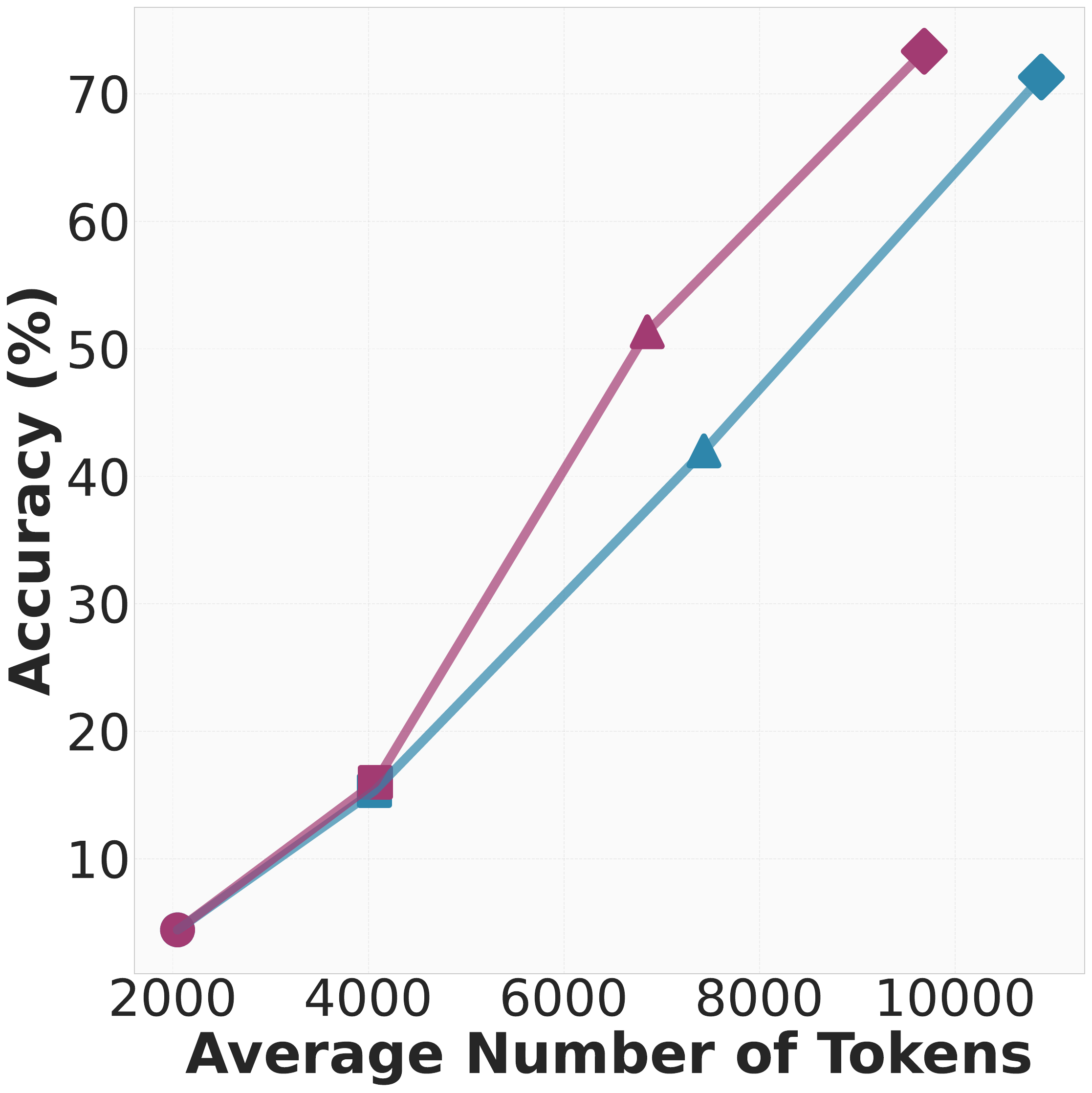}
    \caption{\QwenThreeThirtyTwoB}
    \label{fig:subfig1}
  \end{subfigure}
  \hfill
  \begin{subfigure}[b]{0.2\textwidth}
    \centering
    \includegraphics[width=\textwidth]{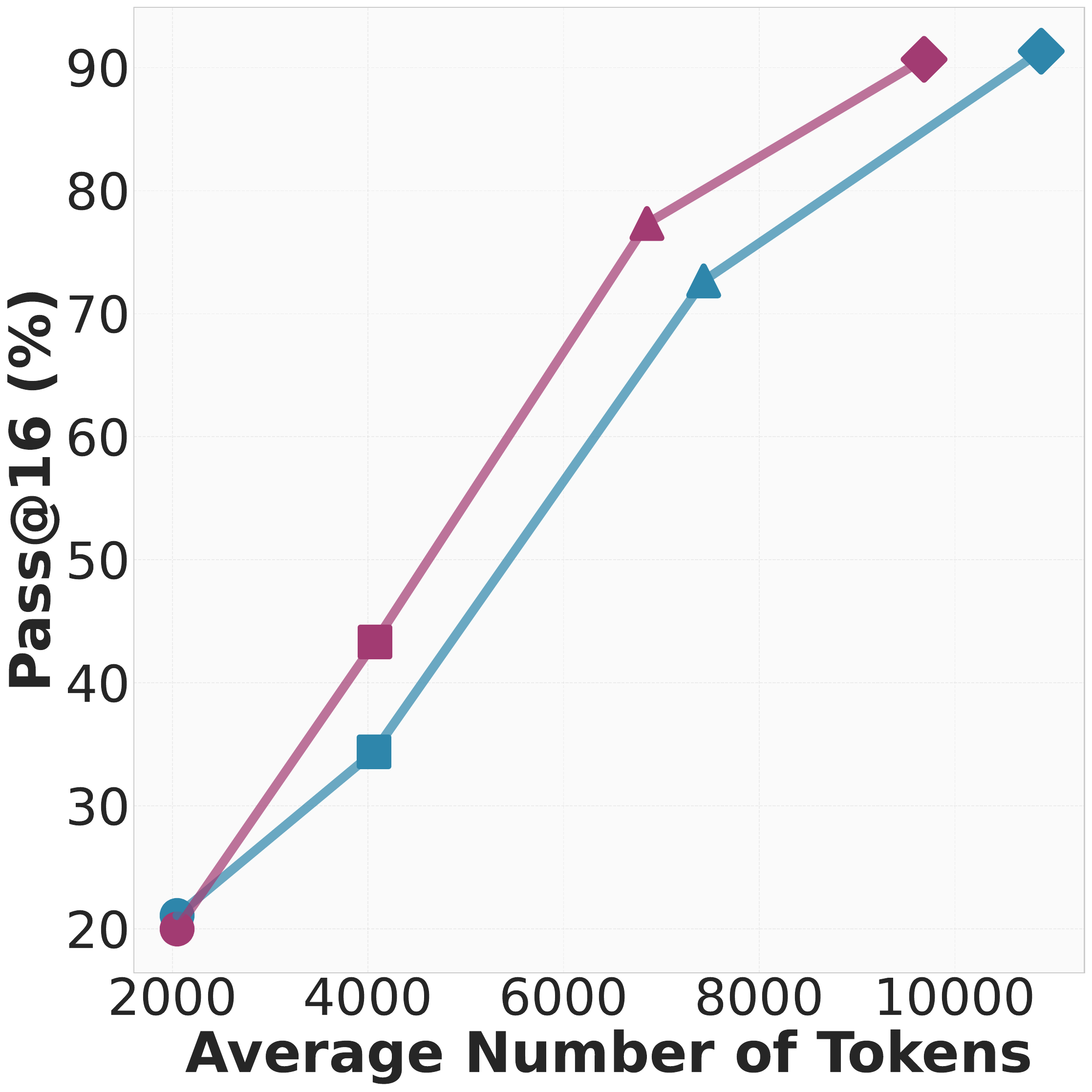}
    \caption{\QwenThreeThirtyTwoB}
    \label{fig:subfig2}
  \end{subfigure}
    \centering
  \includegraphics[width = \textwidth]{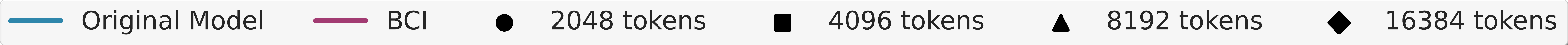}

  \caption{\textbf{Behavior-conditioned Inference (BCI) for AIME\textendash24}. This figure presents results for the AIME\textendash24 dataset. The accuracy is averaged across 80 runs. Pass@16 is averaged across 5 different sets of 16 runs each. The x-axis denotes the average number of tokens generated across all solutions. Each point on the line indicates a given token-budget which is enforced during generation as shown in the legend. The proposed approach improves the token efficiency of the generated solutions achieving superior or competitive performance while producing significantly lesser number of tokens.}
  \label{fig:aime_24_comparison}
\end{figure}

\begin{figure}[htbp]
  \centering
  \begin{subfigure}[b]{0.2\textwidth}
    \centering
    \includegraphics[width=\textwidth]{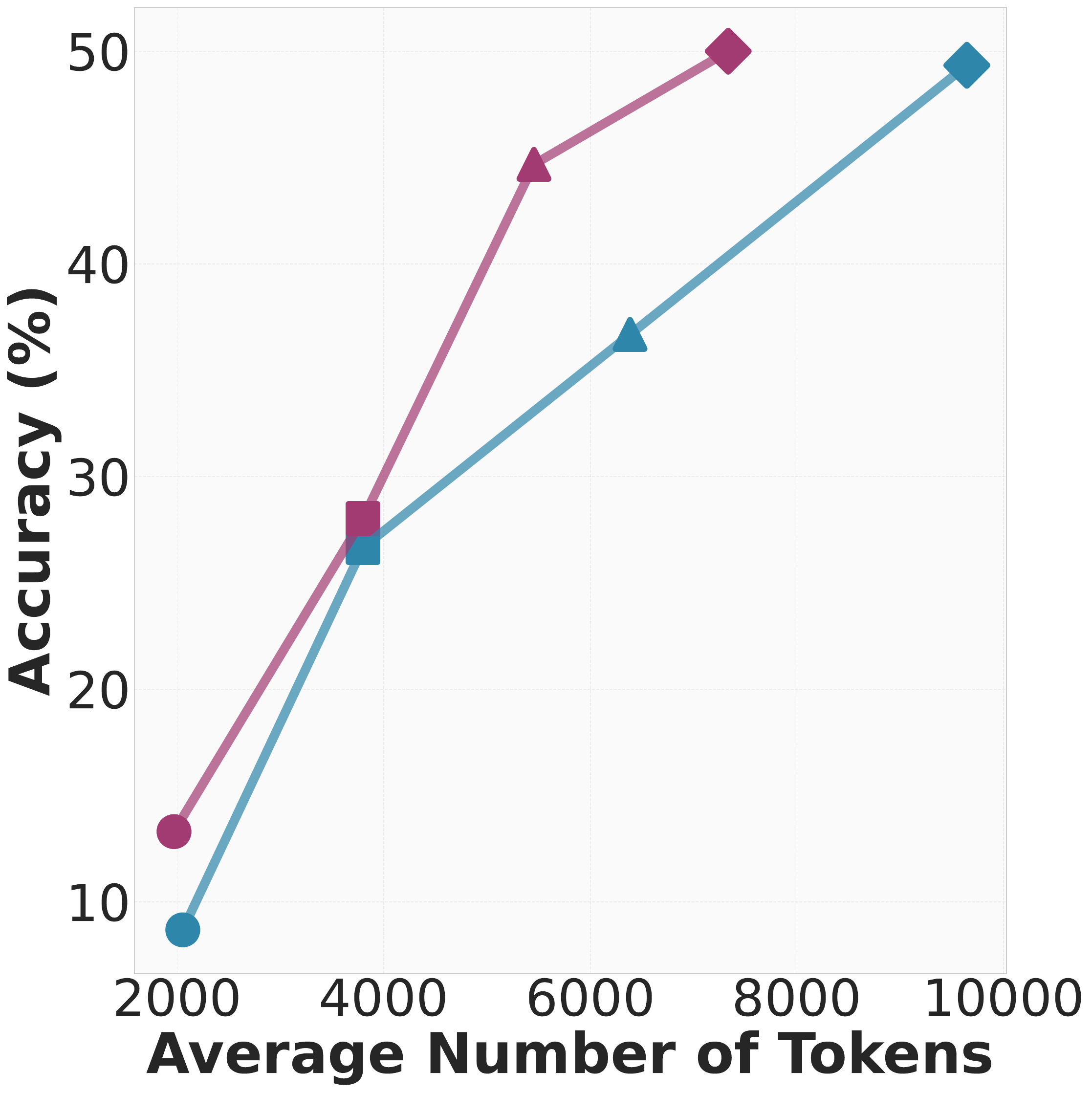}
    \caption{\ROneLlamaSeventyB}
    \label{fig:subfig2}
  \end{subfigure}
   \hfill
  \begin{subfigure}[b]{0.2\textwidth}
    \centering
    \includegraphics[width=\textwidth]{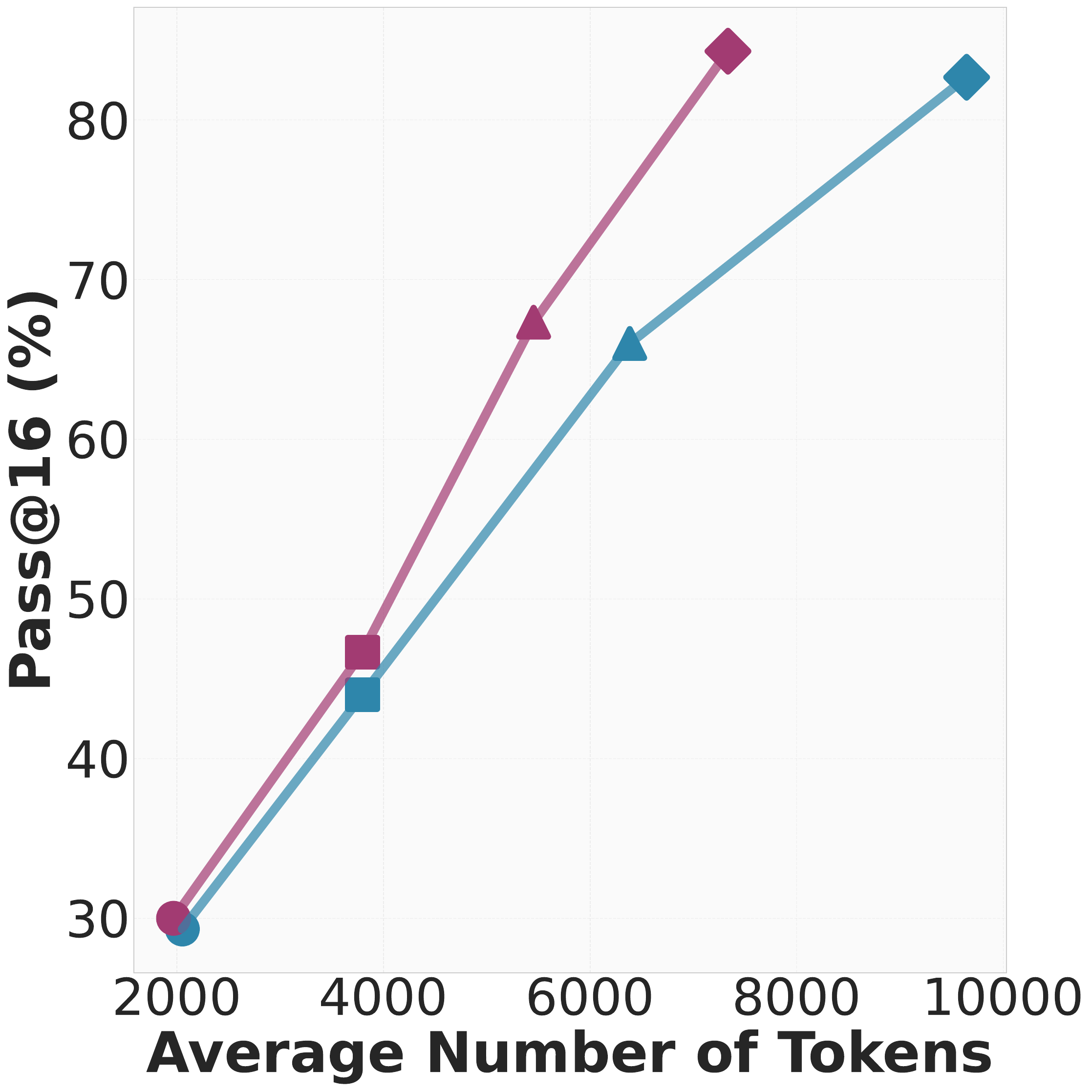}
    \caption{\ROneLlamaSeventyB}
    \label{fig:subfig2}
  \end{subfigure}
    \hfill
  \begin{subfigure}[b]{0.2\textwidth}
    \centering
    \includegraphics[width=\textwidth]{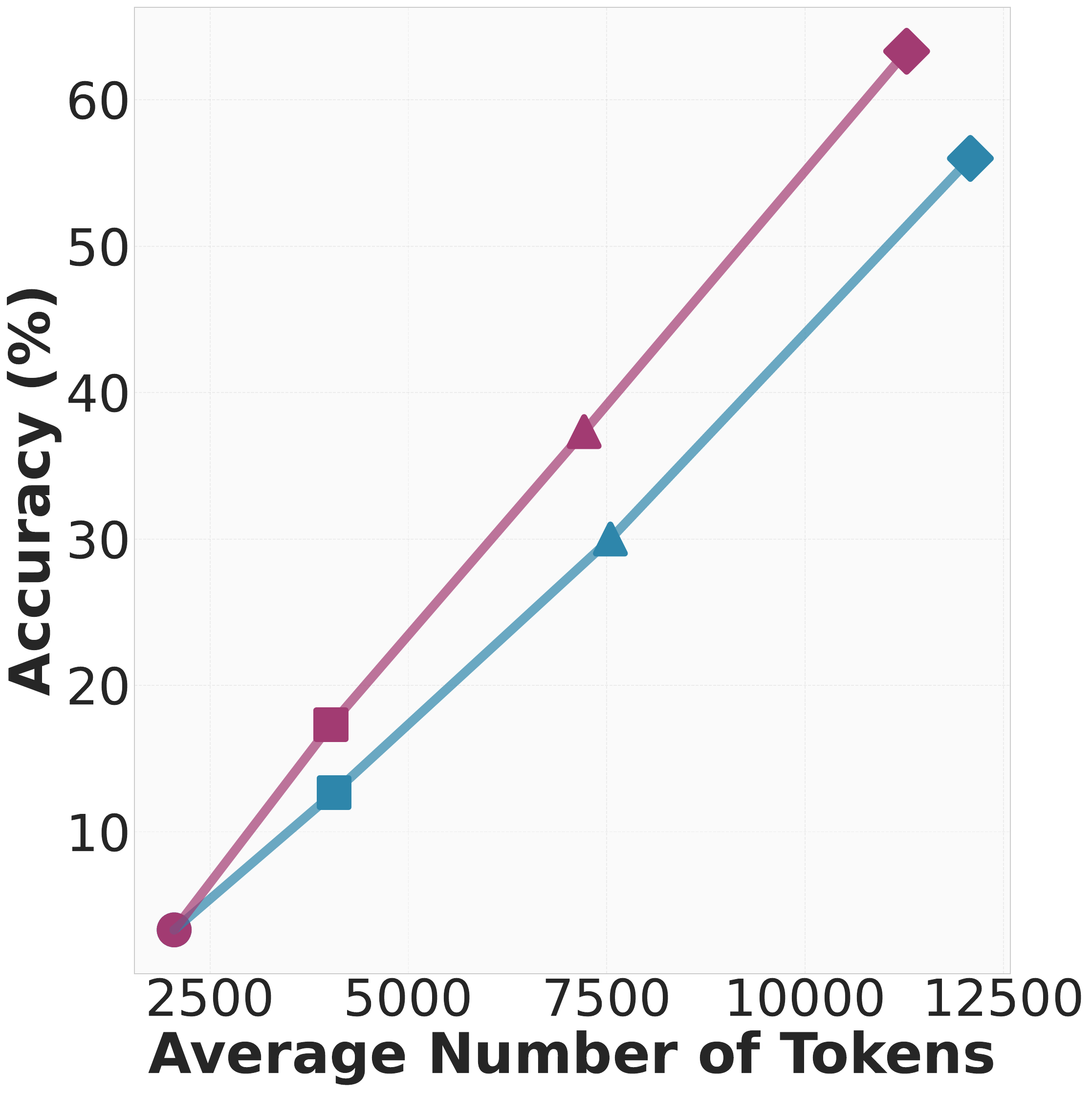}
    \caption{\QwenThreeThirtyTwoB}
    \label{fig:subfig1}
  \end{subfigure}
  \hfill
  \begin{subfigure}[b]{0.2\textwidth}
    \centering
    \includegraphics[width=\textwidth]{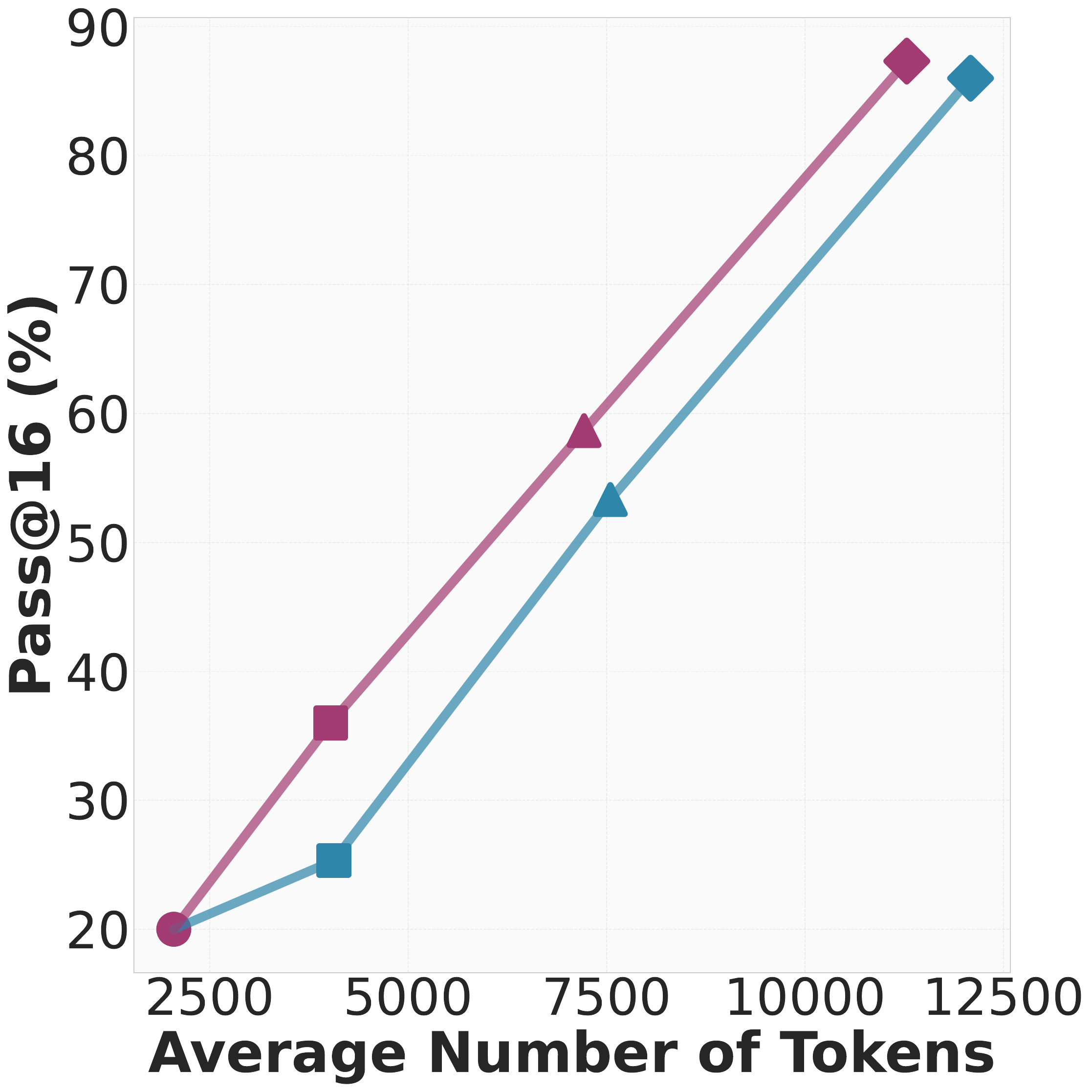}
    \caption{\QwenThreeThirtyTwoB}
    \label{fig:subfig2}
  \end{subfigure}

  \centering
  \includegraphics[width=\textwidth]{figures/aime_24/combined_legend.pdf}

  \caption{\textbf{Behavior-conditioned Inference (BCI) for AIME\textendash25}. This figure presents results for the AIME\textendash25 dataset using behavior list extracted using AIME\textendash22 and 23. The accuracy is averaged across 80 runs. Pass@16 is averaged across 5 different sets of 16 runs each. The x-axis denotes the average number of tokens generated across all solutions. Each point on the line indicates a given token-budget which is enforced during generation. The proposed approach improves the token efficiency of the generated solutions achieving superior or competitive performance while producing significantly lesser number of tokens. }
  \label{fig:aime_25_comparison}
\end{figure}

The results for this experiment are presented in Figures \ref{fig:aime_24_comparison} and \ref{fig:aime_25_comparison}. The accuracy is averaged across 80 runs per question and the pass@16 averaged over 5 sets of 16 responses each. BCI leads to more token efficient solutions achieving improved or competitive performance while using fewer tokens. 

\paragraph{Efficiency Considerations} The proposed BCI approach enables models to produce  fewer output tokens without sacrificing performance and scalability. This reduction in generation length has the potential to substantially lower the cost of inference compared to conventional prompting strategies. While the proposed method involves a larger number of input tokens due to the inclusion of retrieved behaviors, this overhead is mitigated by two key factors. First, the input representations of behaviors can be pre-computed and reused across different queries, amortizing the cost over multiple inferences. Second, there is no autoregressive generation required on the input side, which makes processing these tokens much faster. Moreover, in many proprietary model APIs, input tokens are billed at a lower rate than output tokens, making the proposed approach even more attractive from a cost-efficiency standpoint.

\subsection{Self-Improvement using Behaviors} 
\label{sec:self-improvement}

This section evaluates a model's self-improvement capabilities by using it as the \metacogexpert and the \student at the same time. \ROneLlamaSeventyB is used for both the roles.

\paragraph{Critique-and-Revise Baseline}
 The baseline in this experiment uses an LLM for self-improvement by directly prompting it to critique-and-revise its own past reasoning trace. Concretely, given a question $Q$, the model first produces an initial reasoning trace $R_1$ ($Q \rightarrow R_1$). This reasoning trace along with the original question is then fed back into the model prompting it to generate an improved trace $R_2$ that corrects errors or extends incomplete reasoning ($(Q, R_1) \rightarrow R_2$). This baseline is called the \textit{critique-and-revise} baseline. In this experiment, the token budget for $R_1$ is set to 2{,}048 and that of $R_2$ is varied from 2{,}048 to 16{,}384.

\paragraph{Behavior-guided variant}
In the behavior-guided variant, behaviors are curated from the original reasoning traces $R_1$ which are generated at a token budget of 2{,}048. Following this, each step in the behavior curation pipeline is executed at a token budget of 2{,}048. A total of 16 reasoning traces are generated per question which are used by the \metacogexpert to curate a \behaviorlist for that question. Finally, these behaviors are fed back into the model to generate improved reasoning traces at budgets ranging from 2{,}048 to 16{,}384 ($(B,Q)\!\rightarrow\!R_2$).

\begin{wrapfigure}{r}{0.65\textwidth} 
  \vspace{-\baselineskip} 
  \centering
  \captionsetup{width=0.58\textwidth}

  \begin{subfigure}[b]{0.28\textwidth}
    \centering
    \includegraphics[width=\linewidth]{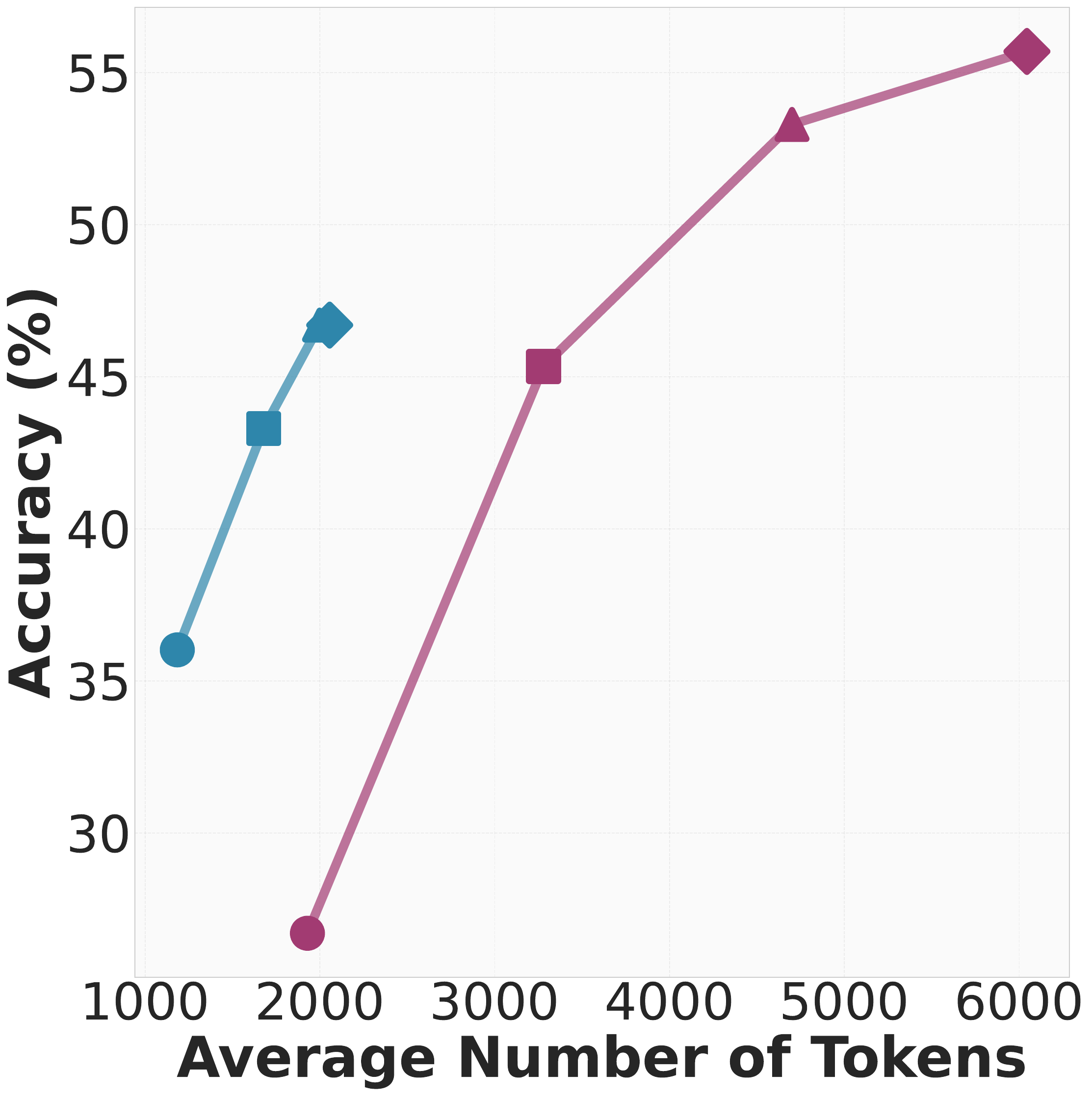}
    \caption{Accuracy on AIME\textendash24}
    \label{fig:wrap_aime24}
  \end{subfigure}\hfill
  \begin{subfigure}[b]{0.28\textwidth}
    \centering
    \includegraphics[width=\linewidth]{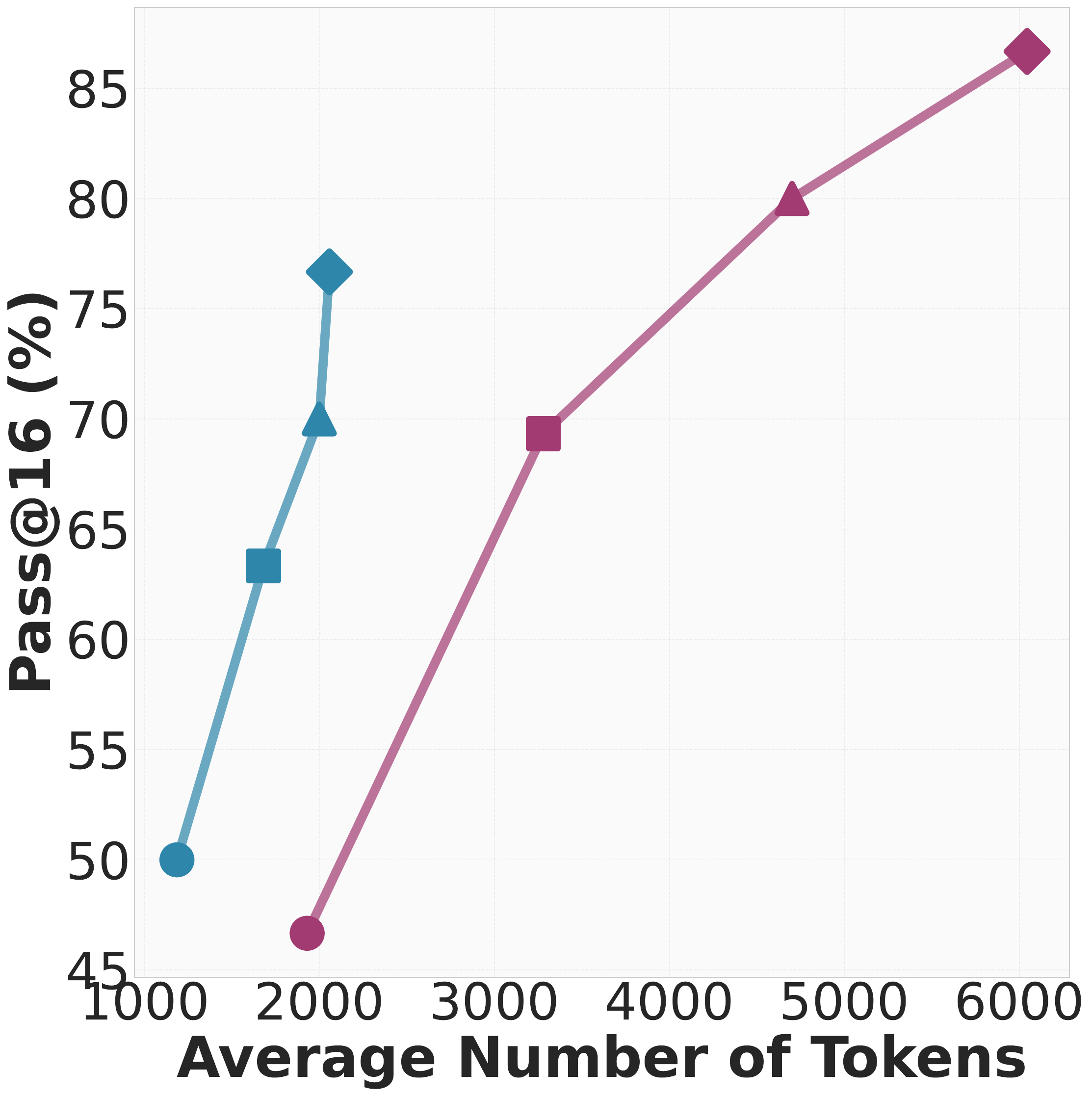}
    \caption{Pass@16 on AIME\textendash24}
    \label{fig:wrap_aime25}
  \end{subfigure}
\centering
  \includegraphics[width=0.60\textwidth]{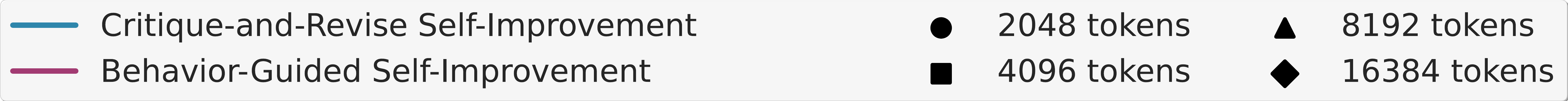}

  \caption{\textbf{Self-improvement} Comparison of the critique-and-revise baseline ($(Q,R_1)\!\rightarrow\!R_2$) with the proposed behavior-guided variant ($(B,Q)\!\rightarrow\!R_2$) on AIME\textendash24. The accuracy is averaged across 80 runs. Pass@16 is averaged across 5 different sets of 16 runs each. The x-axis denotes the average number of tokens generated across all solutions. The behavior-guided approach produces more tokens for a given token-budget compared to the critique-and-revise baseline. However, it also performs significantly better than the critique-and-revise baseline. All runs use \ROneLlamaSeventyB with the initial trace budget fixed at 2{,}048 tokens and the revision budget varied from 2{,}048 to 16{,}384.}
  \label{fig:self_improvement}
\end{wrapfigure}

\paragraph{Results} Results for this experiment are presented in Figure \ref{fig:self_improvement}. Three consistent patterns emerge. (1) Accuracy gains: Conditioning on extracted behaviors ($(B,Q)\!\rightarrow\!R_2$) outperforms the critique-and-revise baseline ($(Q,R_1)\!\rightarrow\!R_2$) at almost every revision token budget evaluated. The gap is modest at low budgets but widens as the available generation budget increases, indicating that the behavior hints help the model make better use of additional tokens. (2) Test time scaling: Performance for behavior-guided self-improvement improves smoothly with increasing budgets thus maintaining the test-time scaling property of the original \ROneLlamaSeventyB model while the critique-and-revise approach struggles to scale performance by utilizing higher token budgets. (3) Token-cost tradeoff: As opposed to the observations from the previous sections, the behavior-guided approach is less token-efficient than the baseline in this case producing more output tokens than the baseline.

\subsection{Supervised fine-tuning with Behaviors} \label{sec:sft}

Behavior-conditioned supervised fine-tuning (BC\textnormal{-}SFT) tries to incorporate good behaviors into the model's parameters. In this setting, \ROneLlamaSeventyB is the \metacogexpert which generates behaviors as well as the \teacher which generates behavior-conditioned responses for training. The following 4 candidates are used as \student models which are fine-tuned: \texttt{Qwen2.5‑14B} \citep{qwen2025qwen25technicalreport}, \texttt{Qwen2.5‑32B‑Instruct} \citep{qwen2025qwen25technicalreport}, \texttt{Qwen3‑14B} \citep{bai2023qwen}, and \texttt{Llama‑3.1‑8B} \citep{dubey2024llama}.

\paragraph{Dataset construction.}

The problems from the S1 dataset \citep{muennighoff2025s1} are used to create the training datasets used in this experiment. For each problem $Q_i$, the \metacogexpert (\ROneLlamaSeventyB) is used to create a set of behaviors using the pipeline from Section \ref{sec:thought_pipeline}. A single reasoning trace $R_i$ is used per problem to create the \behaviorlist. Each step in the pipeline is run at a token budget of 14{,}000. Next, using BCI, a dataset of behavior-conditioned responses is curated using the \teacher (\ROneLlamaSeventyB) model for training the \student models. 

\[
  \mathcal{D}_{\text{BC}}=\{(Q_0,\hat{R}_0),\dots,(Q_n,\hat{R}_n)\}.
\]

where, $\hat{R}_i$ is the behavior-conditioned response generated from the \teacher. The behaviors are retrieved from the \behaviorlist based on question matching i.e. behaviors generated from a given question are used in-context for generating a response to that question. 
For the baseline, the \student models are trained on the corpus of the original reasoning traces generated from the \teacher with normal inference.
\[
  \mathcal{D}_{\text{SFT}}=\{(Q_0,R_0),\dots,(Q_n,R_n)\}.
\]
Importantly, behaviors are \emph{not} provided as input during fine‑tuning on $\mathcal{D}_{\text{BC}}$, or in-context during test-time; any benefit at test time therefore reflects knowledge distilled into the parameters themselves.

\begin{figure}[htbp]
  \centering
  
  \begin{subfigure}[b]{0.23\textwidth}
    \centering
    \includegraphics[width=\textwidth]{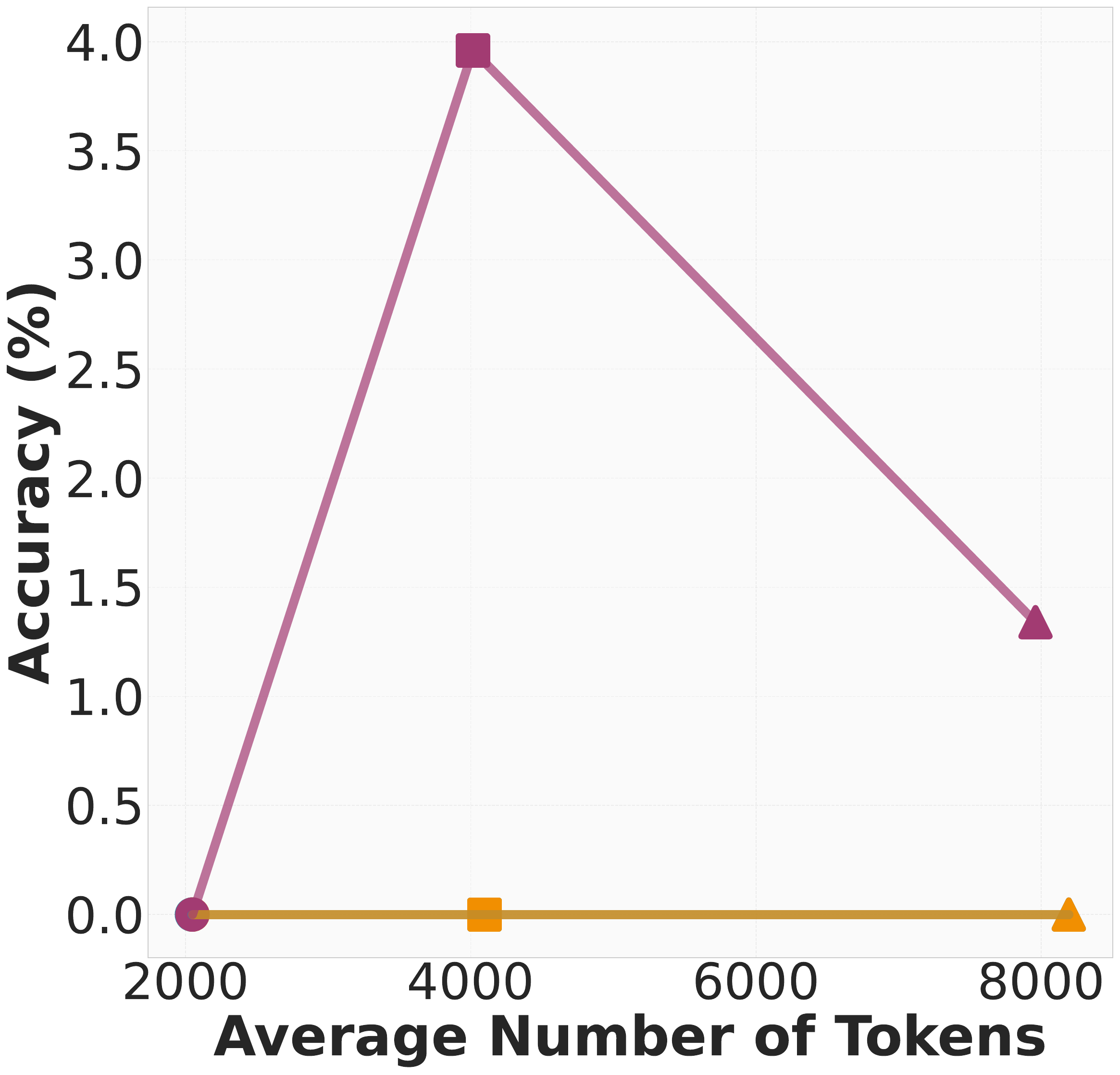}
    \caption{\footnotesize\ttfamily Llama-3.1-8B-Instruct}
    \label{fig:subfig2}
  \end{subfigure}
   \hfill
  \begin{subfigure}[b]{0.23\textwidth}
    \centering
    \includegraphics[width=\textwidth]{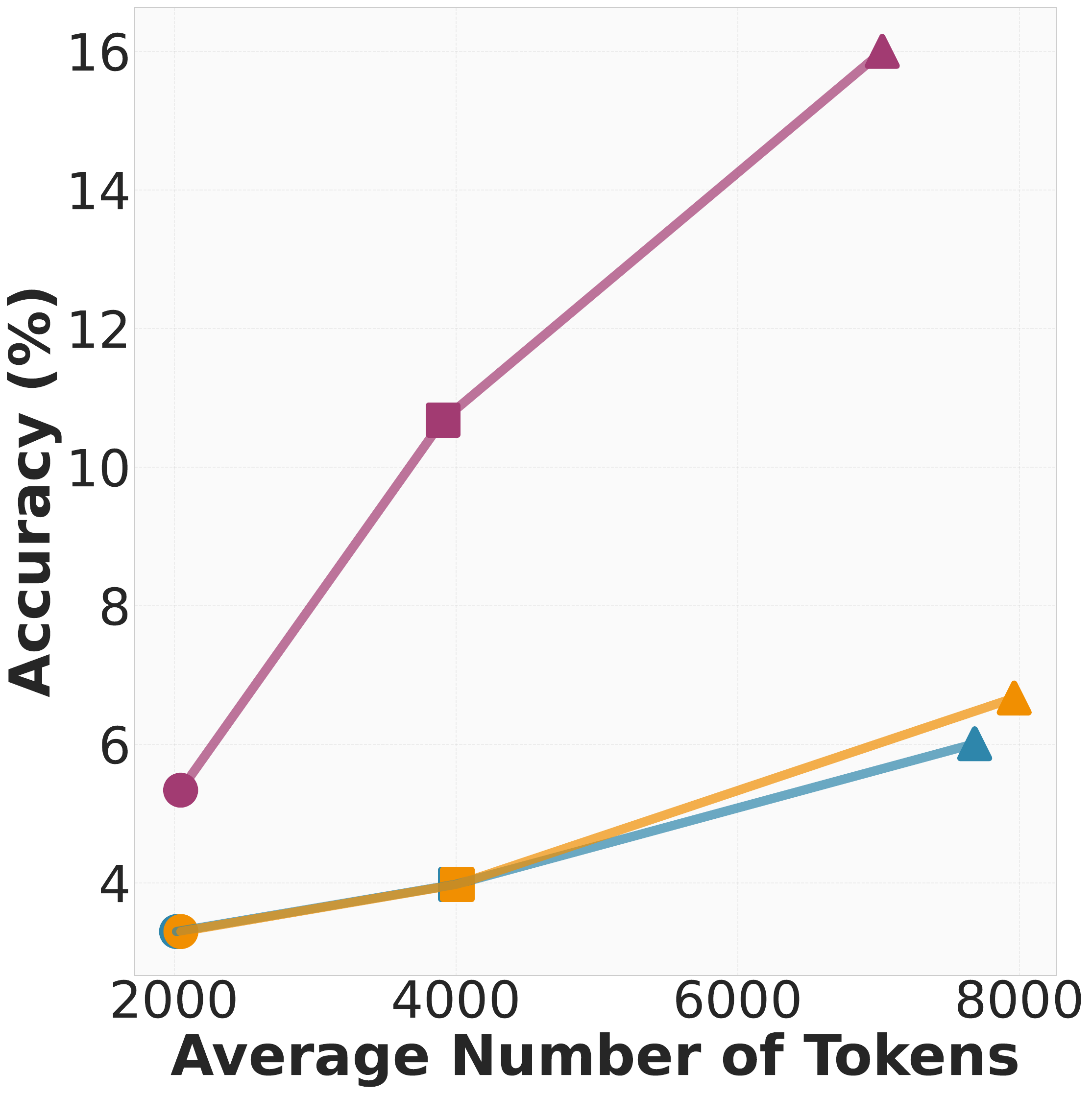}
    \caption{\footnotesize\ttfamily Qwen-2.5-14B-Base}
    \label{fig:subfig2}
  \end{subfigure}
    \hfill
  \begin{subfigure}[b]{0.23\textwidth}
    \centering
    \includegraphics[width=\textwidth]{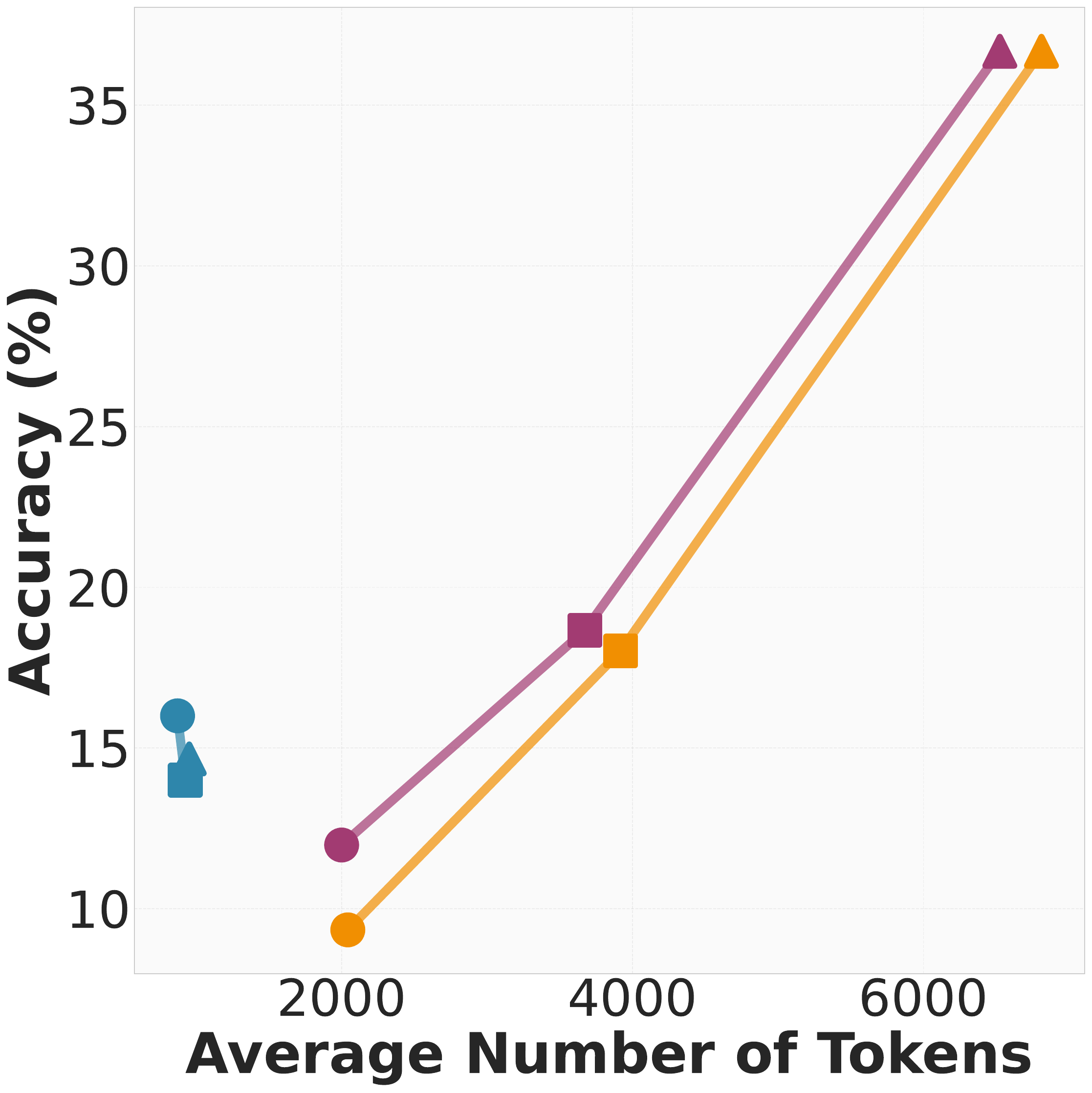}
    \caption{\footnotesize\ttfamily Qwen2.5-32B-Instruct}
    \label{fig:subfig1}
  \end{subfigure}
  \hfill
  \begin{subfigure}[b]{0.23\textwidth}
    \centering
    \includegraphics[width=\textwidth]{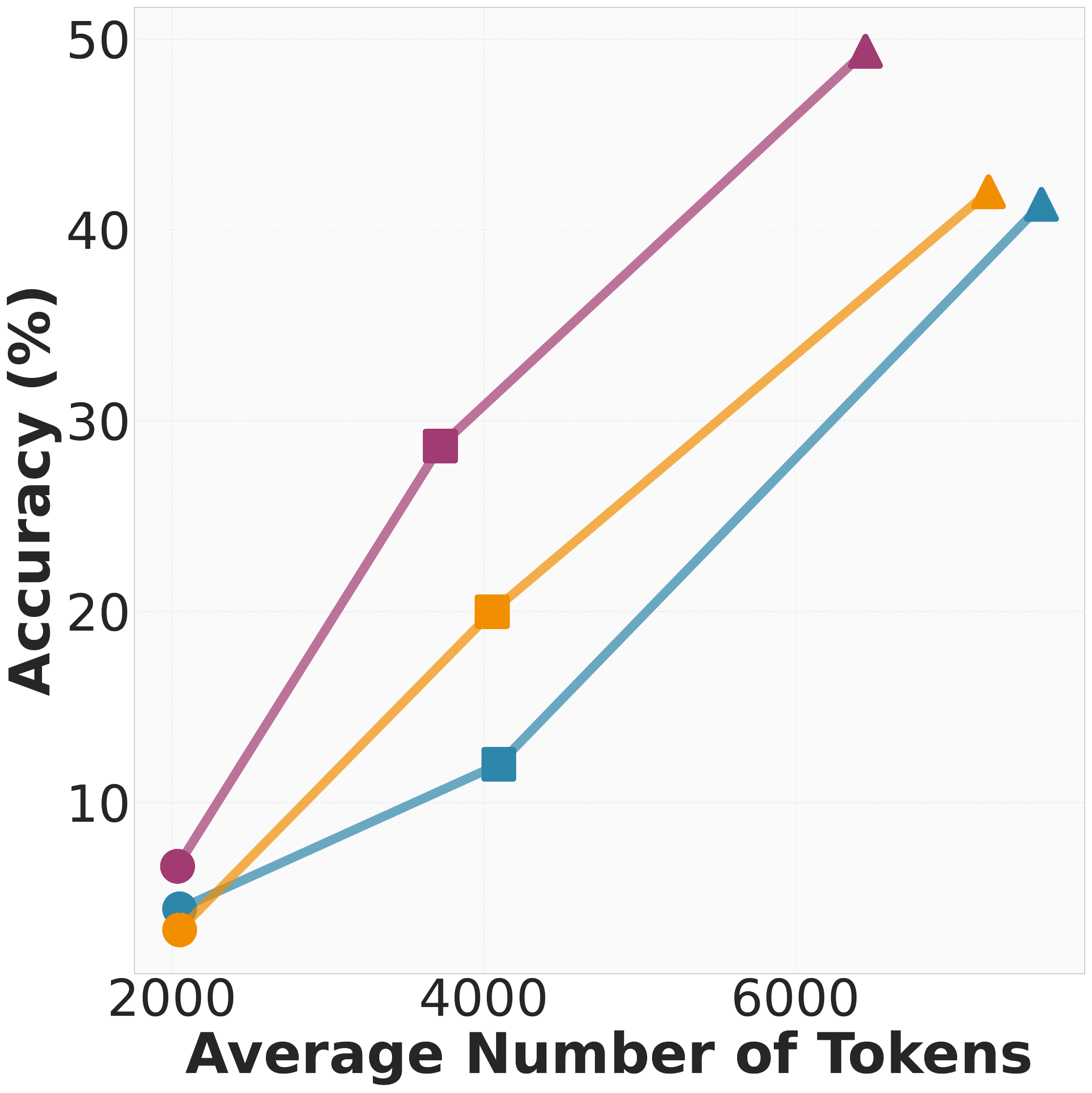}
    \caption{\footnotesize\ttfamily Qwen3-14B}
    \label{fig:subfig2}
  \end{subfigure}

  \vspace{1ex} 

  \centering
  \includegraphics[width=\textwidth]{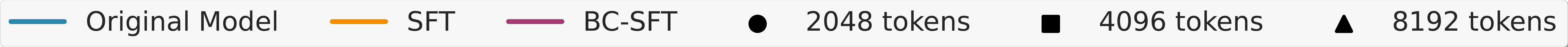}

  \caption{\textbf{SFT Experiment: AIME\textendash24}.
Each panel plots accuracy (\%) versus the average number of generated tokens for one base model. 
Three variants are evaluated: \emph{Original} (no additional training), \emph{SFT} (fine‑tuned on the original corpus $\mathcal{D}_{\text{SFT}}$), and \emph{BC‑SFT} (fine‑tuned on behavior‑conditioned SFT corpus $\mathcal{D}_{\text{BC\textnormal{-}SFT}}$). 
The accuracy is averaged across 80 runs. Each point on the line indicates a given token-budget which is enforced during generation as shown in the legend. The BC\textnormal{-}SFT fine-tuned model consistently achieves superior performance compared to the other two variants across all token budgets while also being more token efficient.}
  \label{fig:sft_aime_24}
\end{figure}

\paragraph{Training setup and evaluation.}
The same setting as that of \citet{muennighoff2025s1} is adapted for this experiment. Each model is fine‑tuned with a 16{,}384‑token context window. At inference only the question is supplied to the model and decoding is performed with budgets from 2{,}048 to 8{,}192 tokens.

\begin{figure}[htbp]
  \centering
  
  \begin{subfigure}[b]{0.23\textwidth}
    \centering
    \includegraphics[width=\textwidth]{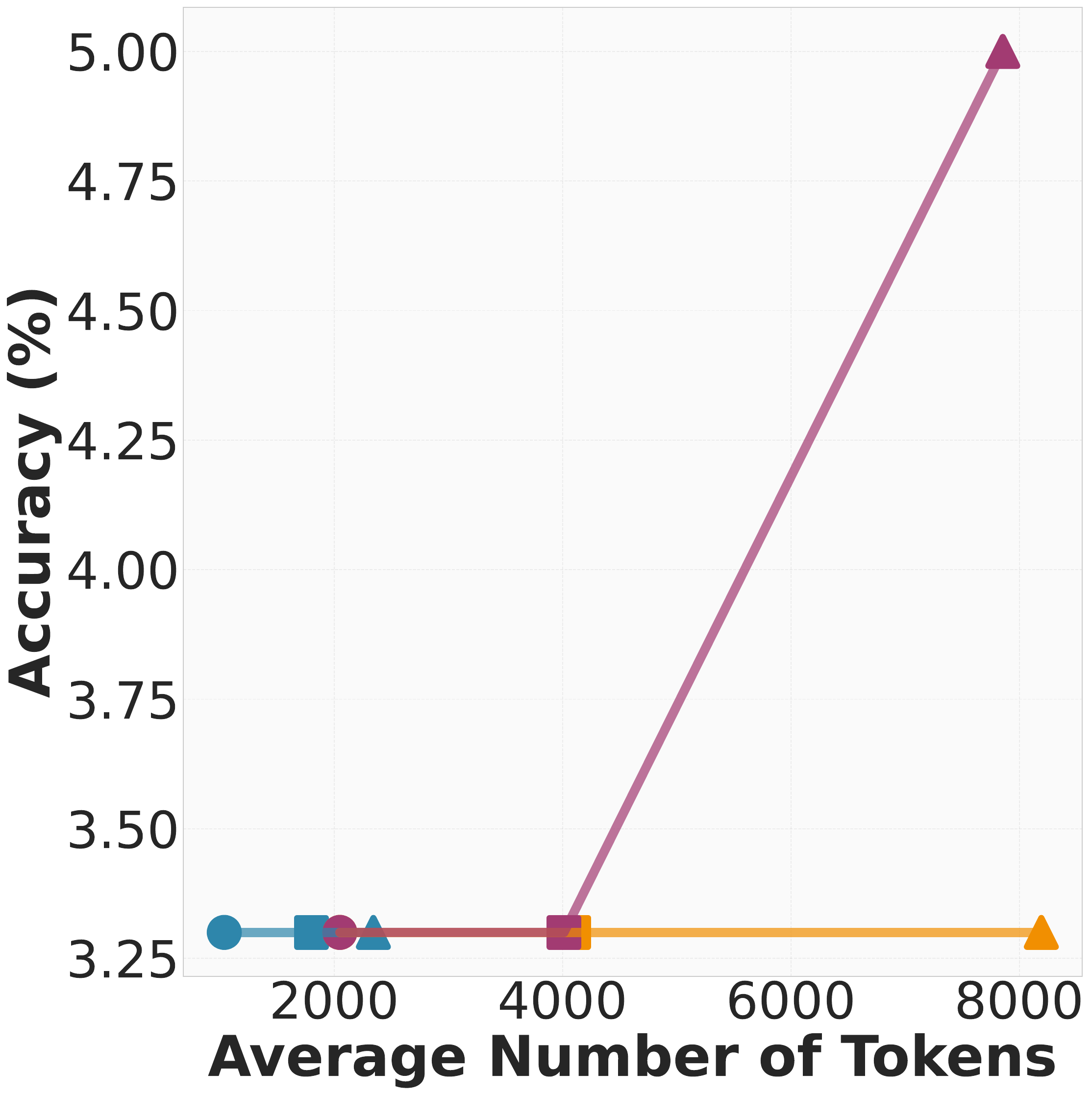}
    \caption{\footnotesize\ttfamily Llama-3.1-8B-Instruct}
    \label{fig:subfig2}
  \end{subfigure}
   \hfill
  \begin{subfigure}[b]{0.23\textwidth}
    \centering
    \includegraphics[width=\textwidth]{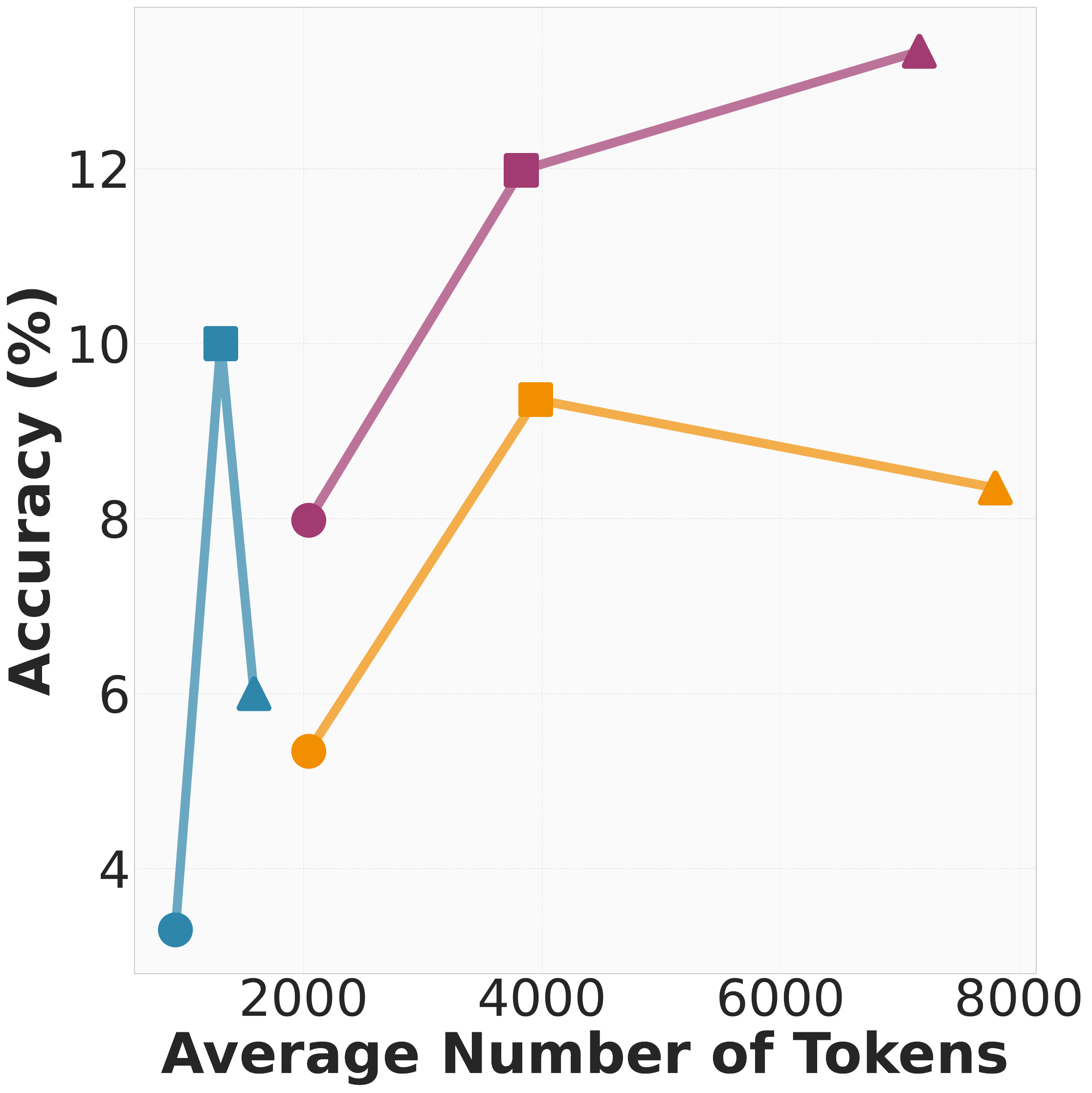}
    \caption{\footnotesize\ttfamily Qwen-2.5-14B-Base}
    \label{fig:subfig2}
  \end{subfigure}
    \hfill
  \begin{subfigure}[b]{0.23\textwidth}
    \centering
    \includegraphics[width=\textwidth]{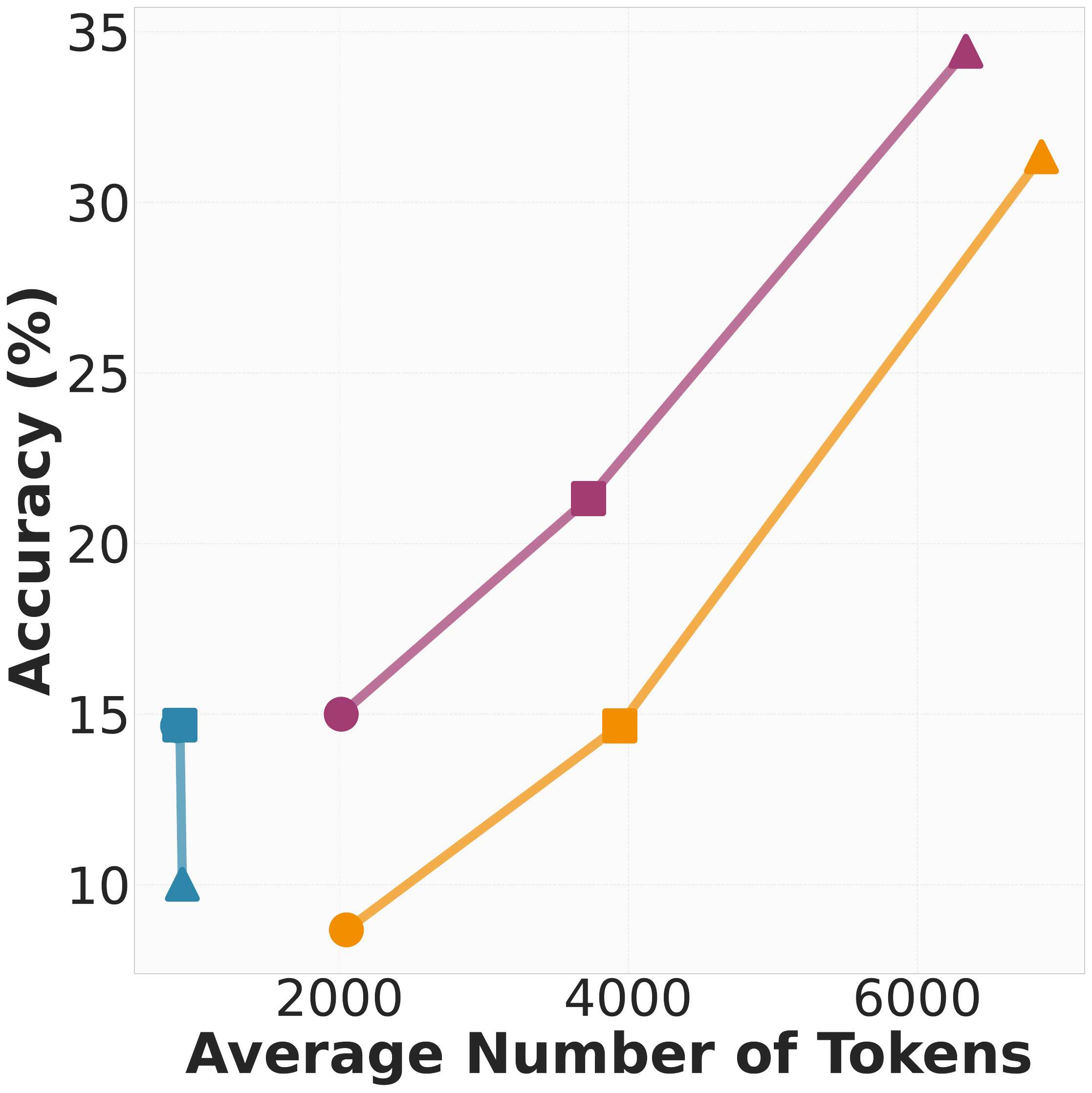}
    \caption{\footnotesize\ttfamily Qwen2.5-32B-Instruct}
    \label{fig:subfig1}
  \end{subfigure}
  \hfill
  \begin{subfigure}[b]{0.23\textwidth}
    \centering
    \includegraphics[width=\textwidth]{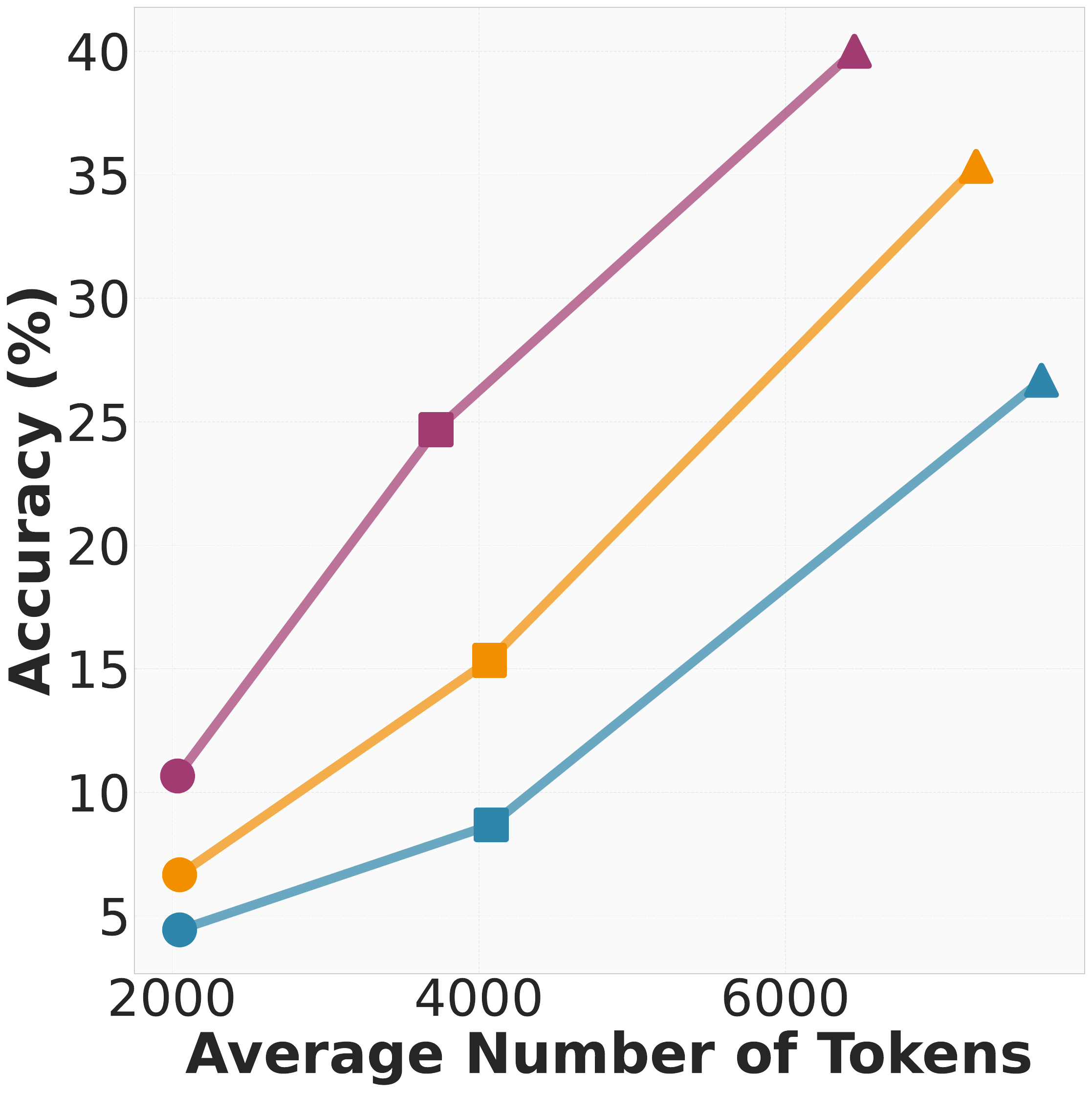}
    \caption{\footnotesize\ttfamily Qwen3-14B}
    \label{fig:subfig2}
  \end{subfigure}

  \vspace{1ex} 

  \centering
  \includegraphics[width=\textwidth]{figures/sft/combined_legend.pdf}

  \caption{\textbf{SFT Experiment: AIME\textendash25}.
Each panel plots accuracy (\%) versus the average number of generated tokens for one base model. 
Three variants are evaluated: \emph{Original} (no additional training), \emph{SFT} (fine‑tuned on the original corpus $\mathcal{D}_{\text{SFT}}$), and \emph{BC‑SFT} (fine‑tuned on behavior‑conditioned SFT corpus $\mathcal{D}_{\text{BC\textnormal{-}SFT}}$). 
The accuracy is averaged across 80 runs. Each point on the line indicates a given token-budget which is enforced during generation as shown in the legend. The BC\textnormal{-}SFT fine-tuned model consistently achieves superior performance compared to the other two variants across all token budgets while also being more token efficient. }
  \label{fig:sft_aime_25}
\end{figure}

\paragraph{Results.}
Figures~\ref{fig:sft_aime_24} and~\ref{fig:sft_aime_25} report AIME\textendash24/25 accuracy as a function of generated tokens for all four base models (\emph{Original}, \emph{SFT}, and \emph{BC\textnormal{-}SFT} variants; 2{,}048--8{,}192 token budgets). 
The results suggest: the BC\textnormal{-}SFT models are not only more token‑efficient, they also deliver \emph{higher} accuracy than both baselines across nearly all budgets. Moreover, BC\textnormal{-}SFT is more effective at transforming non-reasoning models such as \texttt{Qwen2.5\textnormal{-}14B-Base} and \texttt{Qwen2.5\textnormal{-}32B-Instruct} into reasoning models as opposed to the SFT baseline.
This contrasts with the earlier in‑context BCI experiments, where the dominant benefit at large budgets ($>8{,}192$ tokens) was efficiency---models produced fewer tokens while roughly matching (or only slightly exceeding) baseline accuracy. 
Here, by contrast, BC\textnormal{-}SFT confers genuine quality gains: models trained on behavior‑conditioned traces routinely outperform those trained on the original traces even when decoding with the \emph{same} budget, indicating that the fine‑tuning signal imparts useful reasoning behaviors rather than merely teaching the model to be terse. 
To probe whether these gains could be attributed simply to better answer correctness in the training data, the responses in $\mathcal{D}_{\text{SFT}}$ and $\mathcal{D}_{\text{BC}}$ were evaluated against the S1 reference answers, obtaining 42.7\% and 44.4\% accuracy, respectively---a negligible gap that cannot explain the downstream performance deltas. 
Nevertheless, models trained on $\mathcal{D}_{\text{BC}}$ achieve markedly superior AIME performance (see especially panel~(b) in Figures \ref{fig:sft_aime_24} and \ref{fig:sft_aime_25}, \texttt{Qwen2.5\textnormal{-}14B-Base}, where accuracy scales sharply with budget under BC\textnormal{-}SFT while the plain SFT model improves only modestly). 
Taken together, these results strengthen the hypothesis that behavior‑conditioned supervision injects useful intermediate reasoning traits into model parameters, enabling stronger and more efficient problem solving than conventional SFT.

\section{Conclusion and Limitations}

This work introduces a mechanism through which large language models can utilize their metacognitive abilities to distil their own recurring reasoning patterns into concise \textit{behaviors}. Storing and retrieving these behaviors closes a key efficiency gap in LLM reasoning: rather than re‑deriving the same intermediate results, the model simply recalls a relevant behavior and spends its budget on new reasoning. Across three complementary settings—behavior‑conditioned inference, behavior‑guided self‑improvement, and behavior‑conditioned supervised fine‑tuning—the proposed approach demonstrates consistent gains in both accuracy and token efficiency on challenging math benchmarks.

Beyond mathematics, the framework is model‑ and domain‑agnostic, inviting exploration in programming, theorem proving, scientific reasoning, and open‑ended dialogue. Still, several limitations remain. For BCI, the behaviors are retrieved based on the question itself and once they are provided at the beginning, the behavior list is fixed i.e. new behaviors cannot be added to the context. Ideally, a more elegant solution would be that the model retrieves the required behavior on the fly during reasoning. Such, a capability could in-principle be incorporated into the model via training to query the \behaviorlist as a ``tool''. Secondly, the exploration in this paper serves as proof-of-concept to show the benefits of the behavior framework for LLM reasoning, it remains to be seen whether this framework can be scaled to - 1) Curate a large library of behaviors across many domains and retrieve from it during inference; 2) Performing SFT at a larger scale with a massive corpus rewritten using behaviors to improve the smaller models as well as self-improve the model used for curating behaviors and rewriting responses.

Overall, converting slow chains of thought into fast, reusable behaviors enables efficient and scalable reasoning with LLMs, pointing toward LLMs that learn not just to solve problems, but also remember how.

\bibliographystyle{assets/plainnat}
\bibliography{paper}

\clearpage
\newpage
\beginappendix

\input{appendix}

\end{document}

%% file: appendix.tex
\begin{table}[htbp]
\tiny
\centering
\renewcommand{\arraystretch}{1.2}
\begin{tabular}{|l|l|p{7cm}|}
\hline
\textbf{Subject} & \textbf{Behavior Name} & \textbf{Behavior Instruction} \\
\hline

\multirow{5}{*}{Algebra}
& behavior\_use\_systematic\_verification & When performing multi-step calculations, use a systematic approach such as creating a table to track each step's results, ensuring each step is verified before proceeding to the next. \\
& behavior\_simplify\_after\_substitution & After substituting variables, simplify the resulting expression as much as possible to enhance clarity and correctness. \\
& behavior\_recognize\_algebraic\_patterns & Look for common patterns such as perfect squares, cubes, or factorable forms to simplify problem-solving and reduce errors. \\
& behavior\_always\_rationalize\_denominators & Always rationalize denominators to present answers in simplest form, avoiding radicals in the denominator. \\
& behavior\_check\_units & Verify all units are consistent before performing calculations; e.g., ensure time is in years if the rate is annual. \\
\hline

\multirow{5}{*}{Prealgebra}
& behavior\_assign\_variables\_wisely & Use descriptive variable names that match their meaning in the problem to avoid confusion. \\
& behavior\_prioritize\_calculation\_order & Follow the correct order of operations to maintain accuracy. \\
& behavior\_translate\_words\_to\_equation & Convert phrases into math expressions; e.g., "one and one-half of a number" becomes (3/2)x. \\
& behavior\_understand\_independent\_events & Identify independent events and apply the multiplication rule accordingly. \\
& behavior\_check\_for\_zero\_exponents & Confirm non-zero base before applying the rule that anything to the zero power is 1. \\
\hline

\multirow{5}{*}{Counting and Probability}
& behavior\_counting\_principle & Multiply the number of choices for each independent step to get total possibilities. \\
& behavior\_inclusion\_exclusion\_principle & Use \( P(A \cup B) = P(A) + P(B) - P(A \cap B) \) to avoid double-counting. \\
& behavior\_complementary\_probability & Use the complement when it's simpler than calculating direct probability. \\
& behavior\_consider\_symmetry & Account for symmetries in arrangements (e.g., rotations) to reduce unique cases. \\
& behavior\_binomial\_distribution & Use binomial formula \( P(k) = C(n,k)p^k(1-p)^{n-k} \) for fixed, independent binary trials. \\
\hline

\multirow{5}{*}{Geometry}
& behavior\_use\_shoelace\_formula & Use the shoelace formula to find polygon area from coordinates. \\
& behavior\_use\_trigonometric\_relations & Apply Law of Sines, Cosines, or dot products for angle-side relationships. \\
& behavior\_analyze\_quadrant\_signs & Use sign rules for trig functions based on angle quadrant. \\
& behavior\_use\_cross\_product\_for\_area & Use cross product to compute area of polygons using vectors. \\
& behavior\_check\_perpendicularity\_with\_dot\_product & Dot product of zero implies vectors are perpendicular. \\
\hline

\multirow{5}{*}{Precalculus}
& behavior\_apply\_power\_reduction\_formulas & Use power-reduction identities to simplify high powers of sine/cosine. \\
& behavior\_use\_tangent\_addition\_formula & Use \( \tan(A+B) \) formula to simplify expressions involving tangent sums. \\
& behavior\_convert\_to\_polar\_form & Use polar form for complex operations like exponentiation. \\
& behavior\_magnitude\_calculation & Square each component, sum them, and take square root for vector magnitude. \\
& behavior\_check\_orthogonality & Dot product of 0 means vectors are orthogonal. \\
\hline

\multirow{5}{*}{Number Theory}
& behavior\_check\_prime\_factors & Confirm prime factorization is complete and correct for each number. \\
& behavior\_convert\_between\_bases & Expand each digit's place value step-by-step when converting bases. \\
& behavior\_factor\_common\_terms & Factor out shared terms early to simplify later computation. \\
& behavior\_recognize\_repeating\_patterns & Use repeating digit patterns to aid in factoring or GCD steps. \\
& behavior\_systematically\_record\_remainders & Track remainders carefully in Euclidean algorithm or base conversion. \\
\hline

\multirow{5}{*}{Intermediate Algebra}
& behavior\_apply\_vieta\_formulas & Use Vieta's formulas to relate polynomial roots to coefficients. \\
& behavior\_recognize\_telescoping\_patterns & Identify cancellation patterns in fraction products to simplify. \\
& behavior\_use\_calculus & For optimization, derive the function, find critical points using calculus. \\
& behavior\_recognize\_conjugate\_pairs & Know complex roots appear in conjugate pairs in real-coefficient polynomials. \\
& behavior\_triangle\_inequality\_check & Ensure sum of any two sides exceeds the third to form a triangle. \\
\hline

\end{tabular}
\caption{Examples of behaviors curated from the MATH Dataset \citep{hendrycks2021measuring}}
\label{tab:math_behavior}
\end{table}

\begin{table}[htbp]
\footnotesize
\centering
\renewcommand{\arraystretch}{1.2}
\begin{tabular}{|l|p{9cm}|}
\hline
\textbf{Behavior Name} & \textbf{Behavior Instruction} \\
\hline

behavior\_permutation\_with\_restrictions & Use permutation techniques with restrictions to count valid configurations in grid problems. \\
\hline
behavior\_rhombus\_incircle\_distances & In a rhombus, the sum of distances from any point on the incircle to two opposite sides is equal to twice the inradius. \\
\hline
behavior\_opposite\_sides\_sum & In a parallelogram, the sum of distances from any interior point to two opposite sides is constant and equal to the height between those sides. \\
\hline
behavior\_periodicity & Recognize repeating patterns in remainders to efficiently count valid numbers within a range. \\
\hline
behavior\_remainder\_distinctness & Always check that remainders from different moduli are distinct, especially when moduli are not pairwise coprime. \\
\hline
behavior\_dynamic\_programming & Use dynamic programming to break down complex problems into smaller subproblems and solve them recursively. \\
\hline
behavior\_ptolemy\_theorem & Apply Ptolemy's theorem to cyclic quadrilaterals to find relationships between sides and diagonals, such as \(AC \cdot BD = AB \cdot CD + AD \cdot BC\). \\
\hline
behavior\_concurrent\_intersections & For each point where \(k\) lines intersect, adjust the intersection count by subtracting \((\binom{k}{2} - 1)\) to account for overlapping. \\
\hline
behavior\_volume\_ratio\_parallelepiped & Calculate the volume ratio of two noncongruent parallelepipeds by considering the angles between their edges and applying the volume formula for rhombohedrons. \\
\hline
behavior\_vector\_analysis & Use vector methods or barycentric coordinates to simplify geometric problems involving angles and concurrency. \\
\hline
behavior\_fibonacci\_non\_consecutive & Use the Fibonacci sequence to count subsets with no two consecutive elements; for a set of size \(n\), the count is the \((n+2)\)th Fibonacci number. \\
\hline
behavior\_legendre\_formula
 &  Use Legendre's formula to find the exponent of a prime in the factorization of a factorial. \\
\hline
behavior\_modular\_overlap & Use systematic methods like the Chinese Remainder Theorem and inclusion-exclusion to detect and count overlaps in modular conditions. \\
\hline
behavior\_distance\_from\_point\_to\_line & Calculate the distance from a point to a line using \( \frac{|Ax + By + C|}{\sqrt{A^2 + B^2}} \) to verify tangency. \\
\hline
behavior\_perpendicular\_lines & Recognize that radii to tangent points are perpendicular to the tangent line, aiding in finding slopes and equations. \\
\hline
behavior\_pythagorean\_theorem & Use the Pythagorean theorem \(a^2 + b^2 = c^2\) to find the length of a side in a right triangle. \\
\hline
behavior\_diophantine\_simplification & Simplify Diophantine equations by factoring out common terms or recognizing patterns in coefficients. \\
\hline
behavior\_tangent\_properties & Recall that the radius of a circle is perpendicular to the tangent at the point of tangency, aiding in determining the center. \\
\hline
behavior\_mixtilinear\_incircle\_radius & Calculate the radius of a mixtilinear incircle using:
\\ & \(r_A = \frac{R \cdot \sin^2(\frac{A}{2})}{1 - \sin(\frac{A}{2})}\) for ex-mixtilinear, and
\\ & \(r_A = \frac{R \cdot \sin^2(\frac{A}{2})}{1 + \sin(\frac{A}{2})}\) for internal mixtilinear circles. \\
\hline
behavior\_proportion\_analysis & When dealing with proportions before and after an event, set up equations based on initial and final proportions and solve methodically. \\
\hline

\end{tabular}
\caption{Examples of behaviors curated from AIME\textendash22 and AIME\textendash23 sets.}
\label{tab:aime2223_behaviors}
\end{table}

\begin{table}[htbp]
\footnotesize
\centering
\renewcommand{\arraystretch}{1.2}
\begin{tabular}{|l|p{9cm}|}
\hline
\textbf{Behavior Name} & \textbf{Behavior Instruction} \\
\hline

behavior\_grid\_assignment\_counting & Calculate the number of configurations for grid problems by considering \(2^{\text{(rows + columns)}}\), adjusted for constraints. \\
\hline
behavior\_use\_properties\_of\_exponents & Use properties of exponents to combine terms efficiently when simplifying expressions. \\
\hline
behavior\_count\_rectangles\_in\_regular\_polygons & When counting rectangles in a regular polygon, consider perpendicular diagonal pairs and use symmetry to avoid overcounting. For a regular \(n\)-gon, identify all sets of four vertices forming rectangles by checking perpendicularity and equal lengths. \\
\hline
behavior\_domain\_restriction & When dealing with functions like \(\tanh\), which have restricted ranges, always consider their effect on derived variables to set valid domains for optimization. \\
\hline
behavior\_consistent\_signage & Maintain consistent sign conventions, especially when solving geometry problems in coordinate space. \\
\hline
behavior\_convert\_log\_to\_exponential & Convert logarithmic equations to exponential form to simplify variable relationships. \\
\hline
behavior\_transcendental\_equation\_handling & Recognize when equations may not admit algebraic solutions and consider numerical or graphical methods instead. \\
\hline
behavior\_fermats\_little\_theorem & Apply Fermat's Little Theorem: if \(p\) is prime and \(n\) is not divisible by \(p\), then \(n^{p-1} \equiv 1 \mod p\). \\
\hline
behavior\_tangent\_segment\_length & Calculate tangent length from a point to a circle using \(\sqrt{d^2 - r^2}\), where \(d\) is the distance to the center and \(r\) is the radius. \\
\hline
behavior\_time\_conversion & Always convert hours to minutes or vice versa to maintain consistent units throughout time calculations. \\
\hline
behavior\_inradius\_calculation & Use \(r = \text{Area} / s\) to find the inradius of a triangle, where \(s\) is the semi-perimeter. \\
\hline
behavior\_calculus\_optimization & Use calculus to find extrema: compute the derivative, set it to zero, solve for critical points, and verify using the second derivative or sign changes. \\
\hline
behavior\_astroid\_properties & Recognize the astroid formed by envelopes from \((p,0)\) to \((0,q)\) satisfying \(p^2 + q^2 = 1\). The astroid is given by \(x^{2/3} + y^{2/3} = 1\), useful for tangency problems. \\
\hline
behavior\_counting\_symmetric\_pairs & In regular polygons, count figures by analyzing all symmetric vertex pairs and step sizes. For a dodecagon, consider both \(k\) and \(n-k\) step sizes. \\
\hline
behavior\_systematic\_enumeration & Enumerate all cases methodically while honoring constraints to avoid missed or duplicate solutions. \\
\hline
behavior\_apply\_intercept\_theorem & In problems with parallel lines cutting segments, use the intercept theorem to relate segment lengths proportionally. \\
\hline
behavior\_law\_of\_cosines & Use \(c^2 = a^2 + b^2 - 2ab\cos(C)\) to connect side lengths and angles in triangles. \\
\hline
behavior\_mode\_implications & Use the fact that the mode appears more frequently than other values to structure the dataset accordingly. \\
\hline
behavior\_hyperbolic\_identities & Use identities like \(\cosh^2\theta - \sinh^2\theta = 1\) when parametrizing or manipulating hyperbolas. \\
\hline
behavior\_use\_cyclic\_properties & In cyclic quadrilaterals, apply Ptolemy’s Theorem or Power of a Point to relate sides and diagonals. \\
\hline

\end{tabular}
\caption{Example of behaviors obtained from AIME-24 for the self-improvement experiment}
\label{tab:aime24_behaviors}
\end{table}

\begin{table}[htbp]
\footnotesize
\centering
\renewcommand{\arraystretch}{1.2}
\begin{tabular}{|l|p{9cm}|}
\hline
\textbf{Behavior Name} & \textbf{Behavior Instruction} \\
\hline

behavior\_distance\_from\_point\_to\_line & Calculate the distance from a point to a line using \( \frac{|ax + by + c|}{\sqrt{a^2 + b^2}} \). \\
\hline
behavior\_arrhenius\_formula & Apply the Arrhenius formula \(k = A \cdot \exp\left(-\frac{E}{k_B T}\right)\) for temperature-dependent reaction rates. \\
\hline
behavior\_gradient\_check & Verify that the gradient (sum of unit vectors) is zero at a potential minimizer to confirm it's the geometric median. \\
\hline
behavior\_iterative\_methods & Use iterative algorithms like the Weiszfeld method to approximate the geometric median when a closed-form solution is not feasible. \\
\hline
behavior\_dipole\_potential\_inside & For dipole-like surface charge distributions, the electric potential inside a spherical shell behaves like \( r \cos \theta \). \\
\hline
behavior\_permutations\_with\_duplicates & Compute distinct permutations with repeated elements using \( \frac{n!}{n_1! n_2! \cdots n_k!} \). \\
\hline
behavior\_area\_parallelogram\_complex & Compute the area of a parallelogram formed by complex numbers using the imaginary part of \( z_1 \cdot \overline{z_2} \). \\
\hline
behavior\_divergent\_series\_handling & Be cautious with divergent series; sometimes combinations of divergent sums can converge via cancellation. \\
\hline
behavior\_transposition\_cycles & The minimum number of transpositions to sort a permutation is \( n - c \), where \( c \) is the number of cycles. \\
\hline
behavior\_kl\_divergence\_relations & The chi-square statistic is twice the leading term in the Taylor expansion of the Kullback–Leibler divergence. \\
\hline
behavior\_gcd\_usage & Use the GCD to break down integers and solve linear Diophantine equations like \( ax + by = d \). \\
\hline
behavior\_mirror\_formula & Use the mirror formula \( \frac{1}{f} = \frac{1}{v} + \frac{1}{u} \) for image distances in spherical mirrors. \\
\hline
behavior\_kinetic\_energy\_eV & Mean kinetic energy of a gas molecule is \( E = \frac{3}{2} kT \), where \( k = 8.617 \times 10^{-5} \) eV/K. \\
\hline
behavior\_midpoint\_calculation & Find the midpoint of a segment using the average of the endpoints' coordinates. \\
\hline
behavior\_prime\_exponent\_independence & Treat each prime factor independently when handling divisibility conditions; calculate probabilities separately then combine. \\
\hline
behavior\_cauchy\_variation & Add a constant to \( f(x+y) = f(x) + f(y) + C \) and define \( g(x) = f(x) + C \) to convert to a Cauchy equation. \\
\hline
behavior\_lambert\_w\_recognition & Recognize that equations like \( x e^x = k \) can be solved using the Lambert W function: \( x = W(k) \). \\
\hline
behavior\_permutation\_cycles & Consider cycle structure to simplify permutation counting, especially with required or forbidden cycle lengths. \\
\hline
behavior\_cubing\_and\_cube\_roots & Recognize cubes or cube roots to simplify expressions, especially in volume-related problems. \\
\hline
behavior\_lagrange\_multipliers & Use Lagrange multipliers for constrained optimization by forming the Lagrangian and solving the system from partial derivatives. \\
\hline
behavior\_lorentz\_factor\_approximation & Use \( \sqrt{1 - v^2/c^2} \approx 1 - \frac{v^2}{2c^2} \) to approximate time dilation effects at small velocities. \\
\hline

\end{tabular}
\caption{Examples of behaviors curated on the S1 dataset \citep{muennighoff2025s1} for the SFT experiment.}
\label{tab:s1_behaviors}
\end{table}

%% file: ani_paper.bbl
\begin{thebibliography}{36}
\providecommand{\natexlab}[1]{#1}
\providecommand{\url}[1]{\texttt{#1}}
\expandafter\ifx\csname urlstyle\endcsname\relax
  \providecommand{\doi}[1]{doi: #1}\else
  \providecommand{\doi}{doi: \begingroup \urlstyle{rm}\Url}\fi

\bibitem[Bai et~al.(2023)Bai, Bai, Chu, Cui, Dang, Deng, Fan, Ge, Han, Huang, et~al.]{bai2023qwen}
Jinze Bai, Shuai Bai, Yunfei Chu, Zeyu Cui, Kai Dang, Xiaodong Deng, Yang Fan, Wenbin Ge, Yu~Han, Fei Huang, et~al.
\newblock Qwen technical report.
\newblock \emph{arXiv preprint arXiv:2309.16609}, 2023.

\bibitem[Borgeaud et~al.(2022)Borgeaud, Mensch, Hoffmann, Cai, Rutherford, Millican, Van Den~Driessche, Lespiau, Damoc, Clark, et~al.]{borgeaud2022improving}
Sebastian Borgeaud, Arthur Mensch, Jordan Hoffmann, Trevor Cai, Eliza Rutherford, Katie Millican, George~Bm Van Den~Driessche, Jean-Baptiste Lespiau, Bogdan Damoc, Aidan Clark, et~al.
\newblock Improving language models by retrieving from trillions of tokens.
\newblock In \emph{International conference on machine learning}, pages 2206--2240. PMLR, 2022.

\bibitem[Chen et~al.(2024)Chen, Xiao, Zhang, Luo, Lian, and Liu]{chen2024bge}
Jianlv Chen, Shitao Xiao, Peitian Zhang, Kun Luo, Defu Lian, and Zheng Liu.
\newblock Bge m3-embedding: Multi-lingual, multi-functionality, multi-granularity text embeddings through self-knowledge distillation.
\newblock \emph{arXiv preprint arXiv:2402.03216}, 2024.

\bibitem[Didolkar et~al.(2024)Didolkar, Goyal, Ke, Guo, Valko, Lillicrap, Rezende, Bengio, Mozer, and Arora]{didolkar2024metacognition}
Aniket Didolkar, Anirudh Goyal, Nan~Rosemary Ke, Siyuan Guo, Michal Valko, Timothy Lillicrap, Danilo Rezende, Yoshua Bengio, Michael Mozer, and Sanjeev Arora.
\newblock Metacognitive capabilities of llms: An exploration in mathematical problem solving.
\newblock In A.~Globerson, L.~Mackey, D.~Belgrave, A.~Fan, U.~Paquet, J.~Tomczak, and C.~Zhang, editors, \emph{Advances in Neural Information Processing Systems}, volume~37, pages 19783--19812. Curran Associates, Inc., 2024.
\newblock \url{https://proceedings.neurips.cc/paper_files/paper/2024/file/2318d75a06437eaa257737a5cf3ab83c-Paper-Conference.pdf}.

\bibitem[Douze et~al.(2024)Douze, Guzhva, Deng, Johnson, Szilvasy, Mazaré, Lomeli, Hosseini, and Jégou]{douze2024faiss}
Matthijs Douze, Alexandr Guzhva, Chengqi Deng, Jeff Johnson, Gergely Szilvasy, Pierre-Emmanuel Mazaré, Maria Lomeli, Lucas Hosseini, and Hervé Jégou.
\newblock The faiss library.
\newblock 2024.

\bibitem[Dubey et~al.(2024)Dubey, Jauhri, Pandey, Kadian, Al-Dahle, Letman, Mathur, Schelten, Yang, Fan, et~al.]{dubey2024llama}
Abhimanyu Dubey, Abhinav Jauhri, Abhinav Pandey, Abhishek Kadian, Ahmad Al-Dahle, Aiesha Letman, Akhil Mathur, Alan Schelten, Amy Yang, Angela Fan, et~al.
\newblock The llama 3 herd of models.
\newblock \emph{arXiv e-prints}, pages arXiv--2407, 2024.

\bibitem[Flavell(1979)]{flavell1979metacognition}
John Flavell.
\newblock Metacognition and cognitive monitoring: A new area of cognitive-developmental inquiry.
\newblock \emph{American Psychologist}, 34:\penalty0 906--911, 10 1979.
\newblock \doi{10.1037/0003-066X.34.10.906}.

\bibitem[Fu et~al.(2024)Fu, Chen, Zhu, Fu, Dai, Qiao, and Zhang]{fu2024efficiently}
Yichao Fu, Junda Chen, Siqi Zhu, Zheyu Fu, Zhongdongming Dai, Aurick Qiao, and Hao Zhang.
\newblock Efficiently serving llm reasoning programs with certaindex.
\newblock \emph{arXiv preprint arXiv:2412.20993}, 2024.

\bibitem[Gao et~al.(2024)Gao, Xu, Ye, Liu, He, Fu, Mei, Wang, and Wu]{gao2024designing}
Jiaxuan Gao, Shusheng Xu, Wenjie Ye, Weilin Liu, Chuyi He, Wei Fu, Zhiyu Mei, Guangju Wang, and Yi~Wu.
\newblock On designing effective rl reward at training time for llm reasoning.
\newblock \emph{arXiv preprint arXiv:2410.15115}, 2024.

\bibitem[Guo et~al.(2025)Guo, Yang, Zhang, Song, Zhang, Xu, Zhu, Ma, Wang, Bi, et~al.]{guo2025deepseek}
Daya Guo, Dejian Yang, Haowei Zhang, Junxiao Song, Ruoyu Zhang, Runxin Xu, Qihao Zhu, Shirong Ma, Peiyi Wang, Xiao Bi, et~al.
\newblock Deepseek-r1: Incentivizing reasoning capability in llms via reinforcement learning.
\newblock \emph{arXiv preprint arXiv:2501.12948}, 2025.

\bibitem[Guu et~al.(2020)Guu, Lee, Tung, Pasupat, and Chang]{guu2020retrieval}
Kelvin Guu, Kenton Lee, Zora Tung, Panupong Pasupat, and Mingwei Chang.
\newblock Retrieval augmented language model pre-training.
\newblock In \emph{International conference on machine learning}, pages 3929--3938. PMLR, 2020.

\bibitem[He et~al.(2021)He, Neubig, and Berg-Kirkpatrick]{he2021efficient}
Junxian He, Graham Neubig, and Taylor Berg-Kirkpatrick.
\newblock Efficient nearest neighbor language models.
\newblock \emph{arXiv preprint arXiv:2109.04212}, 2021.

\bibitem[He et~al.(2025)He, Panigrahi, Lin, and Arora]{he2025adaptmiadaptiveskillbasedincontext}
Yinghui He, Abhishek Panigrahi, Yong Lin, and Sanjeev Arora.
\newblock Adaptmi: Adaptive skill-based in-context math instruction for small language models, 2025.
\newblock \url{https://arxiv.org/abs/2505.00147}.

\bibitem[Hendrycks et~al.(2021)Hendrycks, Burns, Kadavath, Arora, Basart, Tang, Song, and Steinhardt]{hendrycks2021measuring}
Dan Hendrycks, Collin Burns, Saurav Kadavath, Akul Arora, Steven Basart, Eric Tang, Dawn Song, and Jacob Steinhardt.
\newblock Measuring mathematical problem solving with the math dataset.
\newblock \emph{arXiv preprint arXiv:2103.03874}, 2021.

\bibitem[Kaur et~al.(2025)Kaur, Park, Goyal, and Arora]{kaur2025instructskillmix}
Simran Kaur, Simon Park, Anirudh Goyal, and Sanjeev Arora.
\newblock Instruct-skillmix: A powerful pipeline for {LLM} instruction tuning.
\newblock In \emph{The Thirteenth International Conference on Learning Representations}, 2025.
\newblock \url{https://openreview.net/forum?id=44z7HL4mfX}.

\bibitem[Lambert et~al.(2024)Lambert, Morrison, Pyatkin, Huang, Ivison, Brahman, Miranda, Liu, Dziri, Lyu, et~al.]{lambert2024tulu}
Nathan Lambert, Jacob Morrison, Valentina Pyatkin, Shengyi Huang, Hamish Ivison, Faeze Brahman, Lester James~V Miranda, Alisa Liu, Nouha Dziri, Shane Lyu, et~al.
\newblock Tulu 3: Pushing frontiers in open language model post-training.
\newblock \emph{arXiv preprint arXiv:2411.15124}, 2024.

\bibitem[Lewis et~al.(2020)Lewis, Perez, Piktus, Petroni, Karpukhin, Goyal, K{\"u}ttler, Lewis, Yih, Rockt{\"a}schel, et~al.]{lewis2020retrieval}
Patrick Lewis, Ethan Perez, Aleksandra Piktus, Fabio Petroni, Vladimir Karpukhin, Naman Goyal, Heinrich K{\"u}ttler, Mike Lewis, Wen-tau Yih, Tim Rockt{\"a}schel, et~al.
\newblock Retrieval-augmented generation for knowledge-intensive nlp tasks.
\newblock \emph{Advances in neural information processing systems}, 33:\penalty0 9459--9474, 2020.

\bibitem[MAA(2024)]{MAA_AIME_2024}
MAA.
\newblock American invitational mathematics examination -- aime, February 2024.
\newblock \url{https://maa.org/math-competitions/american-invitational-mathematics-examination-aime}.
\newblock American Invitational Mathematics Examination, February 2024.

\bibitem[MAA(2025)]{MAA_AIME_2025}
MAA.
\newblock American invitational mathematics examination -- aime, February 2025.
\newblock \url{https://maa.org/math-competitions/american-invitational-mathematics-examination-aime}.
\newblock American Invitational Mathematics Examination, February 2025.

\bibitem[Muennighoff et~al.(2025)Muennighoff, Yang, Shi, Li, Fei-Fei, Hajishirzi, Zettlemoyer, Liang, Cand{\`e}s, and Hashimoto]{muennighoff2025s1}
Niklas Muennighoff, Zitong Yang, Weijia Shi, Xiang~Lisa Li, Li~Fei-Fei, Hannaneh Hajishirzi, Luke Zettlemoyer, Percy Liang, Emmanuel Cand{\`e}s, and Tatsunori Hashimoto.
\newblock s1: Simple test-time scaling.
\newblock \emph{arXiv preprint arXiv:2501.19393}, 2025.

\bibitem[Ning et~al.(2023)Ning, Lin, Zhou, Wang, Yang, and Wang]{ning2023skeleton}
Xuefei Ning, Zinan Lin, Zixuan Zhou, Zifu Wang, Huazhong Yang, and Yu~Wang.
\newblock Skeleton-of-thought: Large language models can do parallel decoding.
\newblock \emph{Proceedings ENLSP-III}, 2023.

\bibitem[OpenAI(2024)]{openai2024openaio1card}
OpenAI.
\newblock Openai o1 system card, 2024.
\newblock \url{https://arxiv.org/abs/2412.16720}.

\bibitem[Qwen et~al.(2025)Qwen, :, Yang, Yang, Zhang, Hui, Zheng, Yu, Li, Liu, Huang, Wei, Lin, Yang, Tu, Zhang, Yang, Yang, Zhou, Lin, Dang, Lu, Bao, Yang, Yu, Li, Xue, Zhang, Zhu, Men, Lin, Li, Tang, Xia, Ren, Ren, Fan, Su, Zhang, Wan, Liu, Cui, Zhang, and Qiu]{qwen2025qwen25technicalreport}
Qwen, :, An~Yang, Baosong Yang, Beichen Zhang, Binyuan Hui, Bo~Zheng, Bowen Yu, Chengyuan Li, Dayiheng Liu, Fei Huang, Haoran Wei, Huan Lin, Jian Yang, Jianhong Tu, Jianwei Zhang, Jianxin Yang, Jiaxi Yang, Jingren Zhou, Junyang Lin, Kai Dang, Keming Lu, Keqin Bao, Kexin Yang, Le~Yu, Mei Li, Mingfeng Xue, Pei Zhang, Qin Zhu, Rui Men, Runji Lin, Tianhao Li, Tianyi Tang, Tingyu Xia, Xingzhang Ren, Xuancheng Ren, Yang Fan, Yang Su, Yichang Zhang, Yu~Wan, Yuqiong Liu, Zeyu Cui, Zhenru Zhang, and Zihan Qiu.
\newblock Qwen2.5 technical report, 2025.
\newblock \url{https://arxiv.org/abs/2412.15115}.

\bibitem[Shah et~al.(2025)Shah, Yu, Lyu, Park, Yu, He, Ke, Mozer, Bengio, Arora, and Goyal]{shah2025aiassistedgenerationdifficultmath}
Vedant Shah, Dingli Yu, Kaifeng Lyu, Simon Park, Jiatong Yu, Yinghui He, Nan~Rosemary Ke, Michael Mozer, Yoshua Bengio, Sanjeev Arora, and Anirudh Goyal.
\newblock Ai-assisted generation of difficult math questions, 2025.
\newblock \url{https://arxiv.org/abs/2407.21009}.

\bibitem[Shao et~al.(2024)Shao, Wang, Zhu, Xu, Song, Bi, Zhang, Zhang, Li, Wu, et~al.]{shao2024deepseekmath}
Zhihong Shao, Peiyi Wang, Qihao Zhu, Runxin Xu, Junxiao Song, Xiao Bi, Haowei Zhang, Mingchuan Zhang, YK~Li, Yang Wu, et~al.
\newblock Deepseekmath: Pushing the limits of mathematical reasoning in open language models.
\newblock \emph{arXiv preprint arXiv:2402.03300}, 2024.

\bibitem[Shi et~al.(2023)Shi, Min, Yasunaga, Seo, James, Lewis, Zettlemoyer, and Yih]{shi2023replug}
Weijia Shi, Sewon Min, Michihiro Yasunaga, Minjoon Seo, Rich James, Mike Lewis, Luke Zettlemoyer, and Wen-tau Yih.
\newblock Replug: Retrieval-augmented black-box language models.
\newblock \emph{arXiv preprint arXiv:2301.12652}, 2023.

\bibitem[Song et~al.(2025)Song, Zheng, Li, Yang, Luo, Pan, and Zhang]{song2025fastcurl}
Mingyang Song, Mao Zheng, Zheng Li, Wenjie Yang, Xuan Luo, Yue Pan, and Feng Zhang.
\newblock Fastcurl: Curriculum reinforcement learning with progressive context extension for efficient training r1-like reasoning models.
\newblock \emph{arXiv e-prints}, pages arXiv--2503, 2025.

\bibitem[Team et~al.(2025)Team, Du, Gao, Xing, Jiang, Chen, Li, Xiao, Du, Liao, et~al.]{team2025kimi}
Kimi Team, Angang Du, Bofei Gao, Bowei Xing, Changjiu Jiang, Cheng Chen, Cheng Li, Chenjun Xiao, Chenzhuang Du, Chonghua Liao, et~al.
\newblock Kimi k1. 5: Scaling reinforcement learning with llms.
\newblock \emph{arXiv preprint arXiv:2501.12599}, 2025.

\bibitem[Trivedi et~al.(2022)Trivedi, Balasubramanian, Khot, and Sabharwal]{trivedi2022interleaving}
Harsh Trivedi, Niranjan Balasubramanian, Tushar Khot, and Ashish Sabharwal.
\newblock Interleaving retrieval with chain-of-thought reasoning for knowledge-intensive multi-step questions.
\newblock \emph{arXiv preprint arXiv:2212.10509}, 2022.

\bibitem[Wei et~al.(2022)Wei, Wang, Schuurmans, Bosma, Xia, Chi, Le, Zhou, et~al.]{wei2022chain}
Jason Wei, Xuezhi Wang, Dale Schuurmans, Maarten Bosma, Fei Xia, Ed~Chi, Quoc~V Le, Denny Zhou, et~al.
\newblock Chain-of-thought prompting elicits reasoning in large language models.
\newblock \emph{Advances in neural information processing systems}, 35:\penalty0 24824--24837, 2022.

\bibitem[Willingham et~al.(1989)Willingham, Nissen, and Bullemer]{WillinghamNissenBullemer1989ProceduralKnowledge}
Daniel~B. Willingham, Matthew~J. Nissen, and Penny Bullemer.
\newblock On the development of procedural knowledge.
\newblock \emph{Journal of Experimental Psychology: Learning, Memory, and Cognition}, 15\penalty0 (6):\penalty0 1047--1060, 1989.
\newblock \doi{10.1037/0278-7393.15.6.1047}.

\bibitem[Xia et~al.(2025)Xia, Leong, Wang, Li, and Li]{xia2025tokenskip}
Heming Xia, Chak~Tou Leong, Wenjie Wang, Yongqi Li, and Wenjie Li.
\newblock Tokenskip: Controllable chain-of-thought compression in llms.
\newblock \emph{arXiv preprint arXiv:2502.12067}, 2025.

\bibitem[Yao et~al.(2023)Yao, Zhao, Yu, Du, Shafran, Narasimhan, and Cao]{yao2023react}
Shunyu Yao, Jeffrey Zhao, Dian Yu, Nan Du, Izhak Shafran, Karthik Narasimhan, and Yuan Cao.
\newblock React: Synergizing reasoning and acting in language models.
\newblock In \emph{International Conference on Learning Representations (ICLR)}, 2023.

\bibitem[Ye et~al.(2025)Ye, Huang, Xiao, Chern, Xia, and Liu]{ye2025limo}
Yixin Ye, Zhen Huang, Yang Xiao, Ethan Chern, Shijie Xia, and Pengfei Liu.
\newblock Limo: Less is more for reasoning.
\newblock \emph{arXiv preprint arXiv:2502.03387}, 2025.

\bibitem[Zeng et~al.(2025{\natexlab{a}})Zeng, Huang, Liu, Liu, He, Ma, and He]{zeng2025simplerl}
Weihao Zeng, Yuzhen Huang, Qian Liu, Wei Liu, Keqing He, Zejun Ma, and Junxian He.
\newblock Simplerl-zoo: Investigating and taming zero reinforcement learning for open base models in the wild.
\newblock \emph{arXiv preprint arXiv:2503.18892}, 2025{\natexlab{a}}.

\bibitem[Zeng et~al.(2025{\natexlab{b}})Zeng, Huang, Li, Zhang, and Deng]{zeng2025done}
Zihao Zeng, Xuyao Huang, Boxiu Li, Hao Zhang, and Zhijie Deng.
\newblock Done is better than perfect: Unlocking efficient reasoning by structured multi-turn decomposition.
\newblock \emph{arXiv preprint arXiv:2505.19788}, 2025{\natexlab{b}}.

\end{thebibliography}
